\documentclass[sigconf]{acmart}

\AtBeginDocument{%
  \providecommand\BibTeX{{%
    \normalfont B\kern-0.5em{\scshape i\kern-0.25em b}\kern-0.8em\TeX}}}

\usepackage{hyperref}       
\usepackage{url}            
\usepackage{booktabs}       
\usepackage{amsfonts}       
\usepackage{nicefrac}       
\usepackage{microtype}      
\usepackage{xcolor}         
\usepackage{graphicx}
\usepackage{subcaption}

\usepackage{amssymb}
\usepackage{mathtools}
\usepackage{amsmath}
\usepackage{comment}
\usepackage{url}
\usepackage{natbib}
\usepackage{bm}
\usepackage{wrapfig}

\usepackage[utf8x]{inputenc}
\usepackage{multirow}
\usepackage{algorithm,algorithmic}

\usepackage{makecell}

\DeclareMathOperator*{\argmin}{arg\,min}

\copyrightyear{2022}
\acmYear{2022}
\setcopyright{rightsretained}

\acmConference[KDD '22]{Proceedings of the 28th ACM SIGKDD Conference on Knowledge Discovery and Data Mining}{August 14--18, 2022}{Washington, DC, USA}
\acmBooktitle{Proceedings of the 28th ACM SIGKDD Conference on Knowledge Discovery and Data Mining (KDD '22), August 14--18, 2022, Washington, DC, USA}
\acmDOI{10.1145/XXXXXXX.XXXXXXX}
\acmISBN{978-1-4503-9385-0/22/08}

\usepackage{etoolbox}
\makeatletter
\patchcmd{\maketitle}{\@copyrightpermission}{
   \begin{minipage}{0.3\columnwidth}
     \href{https://creativecommons.org/licenses/by/4.0/}{\includegraphics[width=0.90\textwidth]{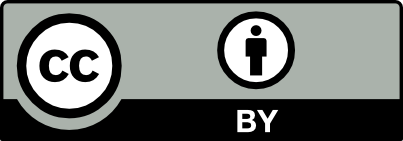}}
   \end{minipage}\hfill
   \begin{minipage}{0.7\columnwidth}
     \href{https://creativecommons.org/licenses/by/4.0/}{This work is licensed under a Creative Commons Attribution International 4.0 License.}
   \end{minipage}

   \vspace{5pt}
}{}{}

\makeatother

\begin{document}

\title{
Multi-Variate Time Series Forecasting on Variable Subsets
}


\author{Jatin Chauhan}
\affiliation{\institution{Google Research}}
\email{chauhanjatin@google.com}

\author{‪Aravindan Raghuveer}
\affiliation{\institution{Google Research}}
\email{araghuveer@google.com}

\author{Rishi Saket}
\affiliation{\institution{Google Research}}
\email{rishisaket@google.com}

\author{Jay Nandy}
\affiliation{\institution{Google Research}}
\email{jnandy@google.com}

\author{Balaraman Ravindran}
\affiliation{\institution{Indian Institute of Technology, Madras}}
\email{ravi@cse.iitm.ac.in}


\renewcommand{\shortauthors}{Jatin Chauhan et al.}


\begin{abstract}
We formulate a new inference task in the domain of multivariate time series forecasting (MTSF), called Variable Subset Forecast (VSF), where only a small subset of the variables is available during inference. Variables are absent during inference because of long-term data loss (eg. sensor failures) or high$\rightarrow$low-resource domain shift between train / test. To the best of our knowledge, robustness of MTSF models in presence of such failures, has not been studied in the literature.   
Through extensive evaluation, we first show that the performance of state of the art methods degrade significantly  in the VSF setting. 
We propose a non-parametric, wrapper technique that can be applied on top any existing forecast models.  Through systematic experiments across 4 datasets and 5 forecast models, we show that our technique is able to recover close to 95\% performance of the models even when only  15\% of the original variables are present. 
  
\end{abstract}



\maketitle

\section{Introduction}
\label{sec:intro}
Multi-Variate Time Series Forecasting (MTSF) continues to receive significant attention from the research community owing to its applicability in many real-world scenarios\cite{han2021dynamic} 
such as traffic forecasting, air quality forecasting, power load forecasting and medical monitoring. We draw attention to two practical scenarios wherein MTSF models require robustness to data scarcity. Specifically, we present MTSF scenarios where at the time of  inference  the model does not have access to all the variables used in training. 

\noindent 
{\bf1. Long Term Variable Data Unavailability:}
In a majority of Multi-Variate Time Series applications, the most common source of  time series data is   via sensors. Each variable in the Multi-Variate Time Series is the output of a sensor. Sensor failures due to reasons like component malfunction, battery outages  are common in   real-world deployments where they are exposed to adverse weather conditions, dust  and so on. As explained in\cite{yick2008wireless} sensor failures can extend for a very long period of time (multiple months in many cases) before the sensor gets replaced. This causes long-term data unavailability for variables that were originally produced by the then-failed sensors. 

\begin{figure}[t]
	\centering
	\includegraphics[ width=75mm, height=45mm]{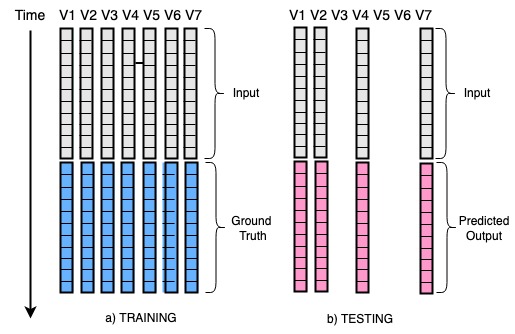}
	
	\caption{Variable Subset Forecast Problem: The two sub-figures show a (i) training and (ii) test instance, respectively. The light gray part on top of each sub-figure represents the input timesteps. The dark shaded part is the forecast. 
	During training, Variables V1 through V7 are present while during Testing, only  V1, V2, V4, V7 are present.  } 
	\label{fig:problem_statement}
\end{figure}

\noindent 
{\bf2. High$\rightarrow$Low Resource Domain Shift:}
Time Series models are  often used in domains that have a wide variability in resource availability. For instance,  consider  inventory prediction of products. This has been posed as a MTSF problem since there exists correlation in market demand across different products (for instance, phones and their cases~\cite{karb2020network}). A model trained on a dataset from large retailer will have a wide variety of products (variables). When applying the same model to a small and medium business(SMB) retailer, the number of products stocked will be significantly limited  and  may also vary over time. So it is not possible to  train a model on a fixed subset of products as the stocked items will vary across SMBs. 
Another example is a  MTSF model to forecast  physiological indicators like blood glucose, insulin, creatinine, etc\cite{cai2020time}. A model trained with data collected at a large hospital would have access to all the variables during training. When the same model is applied (at inference) in a rural hospital without access to many of the diagnostic instruments a large chunk of variables will be missing. In this case, though it is possible to know the subset of variables apriori, it is not scalable to create and maintain one model per known subset. Furthermore, the algorithm proposed in our work (section \ref{sec:our_method}) is able to outperform or match 
a model that is trained on only a subset of the variables known apriori. Similar situations also occur in other domains such as   battery usage prediction between top-end  (high resource) and low-spec smartphones; and
 wind speed forecasting between coastal (high resource) and deep-sea (low resource) naval stations\cite{hamid2018wind}. 


Multi-Variate Time Series Imputation techniques\cite{luo2018multivariate, tang2020joint} are a relevant area of active research. Imputation methods use global inter-variable patterns  and local variable information  (average, recent values)   to generate plausible missing values. Imputation techniques significantly rely on temporal locality therefore are not very effective when the data is  missing for a  long time duration.  In the {\em High$\rightarrow$Low Resource Domain Shift} scenario, variables are missing entirely.  In this case, imputation methods are even less effective. 
To the best of our knowledge, robustness of MTSF models in presence of such failures, that cause a subset of variables to be completely absent during inference, has not been studied in the literature. Although for completeness, we also provide a comparison to Imputation methods in section \ref{sec:comparison_imputation}.


Conventionally, the MTSF problem aims to simultaneously predict the future values of  $\mathcal{N}$ variables  given their past time series values by accurately modeling inter and intra variable dependencies.  In this paper,
the training setup is unchanged from the standard setting wherein we  assume 
that the training algorithm has access to  all variables (denoted by $\mathcal{N}$). But during inference, the past data is only provided for an arbitrary small subset of variables (denoted by $\mathcal{S}$) and we aim to predict future values of variables in $\mathcal{S}$. We call this the {\em Variable Subset Forecast} (VSF in short) problem in MTSF. Figure \ref{fig:problem_statement} summarizes the problem with a simple example.  


 
 The VSF problem  poses 2 key challenges. First, 
 since a substantial portion of the data ($\mathcal{N-S}$ variables) 
 is absent during inference, it is challenging to recover the loss incurred compared to the setting where all $\mathcal{N}$ are present. 
 Second, with only a small  $\mathcal{S}$ presented randomly during inference, it is not feasible to take advantage of the rich inter-variable associations in $\mathcal{N}$. Prior works \cite{wu2020connecting,Guo_Lin_Feng_Song_Wan_2019,10.1145/3209978.3210006,DBLP:journals/corr/abs-1803-01271} have shown significant performance gains by 
 exploiting such inter-variable dependencies.
 Also, retraining the model for every subset $\mathcal{S}$ is not practically feasible since $\mathcal{S}$ is not known during training.
 
 In this paper, we also propose a novel solution to improve the performance of the forecasts models and show that it can even recover close to $95\%$ of the best-case performance even with only $15\%$ of the variables available. The proposed algorithm is a {\em wrapper method} in that it can be implemented over any existing forecast model. We take a non-parametric approach to addressing the problem by retrieving nearest neighbors using only the $\mathcal{S}$ variables and using these neighbors to fill missing values. However, this retrieval is inherently biased because neighbors are retrieved using a distance measure in partial dimensions and is therefore different from the traditional k-NN setting. Achieving near-optimal performance with a biased retrieval mechanism is another technical challenge we address. We propose a novel ensemble weighting method to score candidates from the biased retrieval. 
 Our method is simple to implement and more importantly the underlying forecast models do not need re-training. 
 We make the following contributions in this paper:

\begin{enumerate}
\item We formulate a new inference task called {\em Variable Subset Forecast} in MTSF motivated by  failure situations that arise in real-world settings. To the best of our knowledge we are the first to propose this problem formulation.  (Section~\ref{sec:the_problem})

\item We propose a novel {\em wrapper} solution to improve the performance of various models under this setup. The algorithm is very simple to code\footnote{Code available at  \href{https://github.com/google/vsf-time-series}{https://github.com/google/vsf-time-series}} and is agnostic to the choice of the forecast model. (Section~\ref{sec:our_method})

\item We conduct extensive experiments to understand why current forecast models significantly underperform in the VSF setting (Section~\ref{sec:performance_gap}). Then we quantitatively and qualitatively study our proposed solution with thorough ablations. 


\end{enumerate}

\section{Related Work}
MTSF has a rich history with a diverse category of forecast methods proposed over decades. Some of the earliest models were based on \emph{Statistical Modelling}, which include: Vector Auto Regressive (VAR), Vector Auto Regressive Moving Average model (VARMA) \cite{Ltkepohl2005} as well as Gaussian Process (GP) based methods \cite{GP_time_series}. Another body of literature, although not extensive, is based on leveraging the Matrix Factorization (MF) techniques, which is detailed well in \cite{NIPS2016_85422afb}.
With the surge of Deep Neural models, there has been a series of architectural advances in this domain, which segregates the models into two typical components: (i) first one learns the underlying associations amongst the variables (referred as spatial module) (ii) the second one learns the temporal patterns for the variables' historical data (referred as temporal module). 

The popular works of \cite{10.1145/3209978.3210006, DBLP:journals/corr/abs-1809-04206} provided some of the first LSTM based multivariate forecasting models, followed by more sophisticated models such as DeepState \cite{NEURIPS2018_5cf68969} and subsequently Transformers \cite{10.1145/3447548.3467401}. A huge bottleneck of these works is that interaction amongst the variables is encapsulated in a global hidden vector which fails to explicitly model the pairwise dependencies and interactions. To tackle this, GNN-based models have become the SOTA in recent years. Here, the temporal dependencies are usually modelled by RNN whereas a GNN is applied on the spatial axis. This general framework was initially proposed in traffic prediction models such as DCRNN \cite{li2018diffusion} as well as for skeleton-based action recognition \cite{Yan_Xiong_Lin_2018} but was later adopted for multivariate time series forecasting as well \cite{Guo_Lin_Feng_Song_Wan_2019}.

For our experiments,  we select the following $5$ widely used forecast models 
as a representation of the wide variety of recent techniques in literature -  \textbf{(i) MTGNN} \cite{wu2020connecting}:  This method extracts uni-directed relations among the time series variables via a graph learning module and then performs spatio-temporal modelling, where the temporal module is a temporal convolution network. \textbf{(ii) ASTGCN} \cite{Guo_Lin_Feng_Song_Wan_2019}: This method utilizes a spatial-temporal attention mechanism to learn the dynamic spatial and temporal correlations of the data and then leverages a standard spatio-temporal modelling mechanism for forecasting. \textbf{(iii) MSTGCN} \cite{Guo_Lin_Feng_Song_Wan_2019}: This method utilizes graph convolutional networks for capturing the spatial features and convolutions in the temporal dimension for describing dependencies from nearby time slices, to learn to forecast. \textbf{(iv) LSTNet} \cite{10.1145/3209978.3210006}: This model uses convolutions on the spatial axis to discover the associations amongst the variables while a recurrent network is applied on the temporal axis. \textbf{(v) TCN} \cite{DBLP:journals/corr/abs-1803-01271}: This is a completely convolutional neural network based model that combines various practices such as residual connections and dilations with causal convolutions and perform predictions autoregressively.

\section{Problem Formulation}
\label{sec:the_problem}
Consider $N$ distinct variables, each of which represents values for some event that evolves temporally. In the general setting, we have a $P$ timestep input as $ X = \{\bm{Z_{t_1}}, \bm{Z_{t_2}} ... \bm{Z_{t_P}}\}$, where each of the $\bm{Z_{t_i}}$ is a matrix in $\mathbb{R}^{N \times D}$, $D$ being the feature dimension of all the variables at each time step. The first column of $\bm{Z_{t_i}}$ contains the primary value for the variables and the remaining $D-1$ columns represent some auxiliary information \cite{wu2020connecting}. Given a sequence of historical $P$ timesteps, we are required to predict the future values $ Y = \{\bm{z_{t_{P+1}}}, \bm{z_{t_{P+2}}} ... \bm{z_{t_{P+Q}}}\}$, where each $\bm{z_{t_j}}$ is a vector in $\mathbb{R}^{N}$.  $Q$ denotes the forecast horizon,  the number of time steps into the future that we want to predict.  Our aim is to learn a mapping $f : X \rightarrow Y$, also referred to as forecast model.  
During the inference phase, we have a much smaller subset $\mathcal{S}$ of $\mathcal{N}$, i.e., $|\mathcal{S}| \ll |\mathcal{N}|$, where $\mathcal{N}$ is the set of first $N$ integers, $[\![1...N]\!]$. Therefore, each $\bm{Z_{t_i}}$ in the input is now a matrix in $\mathbb{R}^{|\mathcal{S}| \times D}$ during inference. Another important aspect of the problem is that $\mathcal{S}$ is chosen completely at random which makes this setup extremely challenging.

\begin{table*}[h]
\small
\caption{Values of the error metrics for \emph{Partial} and \emph{Oracle} cases. The numbers represent average over the $1000$ selections of $\mathcal{S}$ (the numbers in parenthesis are corresponding standard deviations). \emph{Value} represents the actual errors followed by corresponding $\Delta_{partial}$ w.r.t \emph{Oracle}.} 
\label{table:oracle_vs_subset_setting}

\centering
\setlength\tabcolsep{3.5pt}
\begin{tabular}{|c|l|c|ccc|ccc|ccc|cc|}
\hline
Forecast Model &  Variant  &   &  \multicolumn{3}{c}{\bf \emph{METR-LA}}  &  \multicolumn{3}{c}{\bf \emph{SOLAR}} &  \multicolumn{3}{c}{\bf \emph{Traffic}} & \multicolumn{2}{c|}{\bf \emph{ECG5000}} \\
\cline{4-6}   \cline{7-9} \cline{10-12} \cline{13-14}
& & &  MAE  & RMSE  &  &  MAE  & RMSE  &  &  MAE  & RMSE &  & MAE & RMSE \\

\hline
\multirow{3}{*}{MTGNN} & \multirow{2}{*}{\emph{Partial}} & \emph{Value} & 4.54(0.37) & 8.90(0.68) & & 4.26(0.53) & 6.04(0.81) & & 18.57(2.31) & 38.46(3.94) &   &  3.88(0.61)  & 6.54(1.10)  \\
 & & $\Delta_{partial}$ & 30.08 \% & 23.43 \% & & 44.89 \% & 29.61 \% & & 62.18 \% & 39.95 \%  &   & 13.11 \% & 10.10 \%  \\
 \cline{2-14}
& \emph{Oracle} &  & 3.49(0.25) & 7.21(0.50) & & 2.94(0.27) & 4.66(0.57) & & 11.45(0.57) & 27.48(2.14) &  & 3.43(0.54) & 5.94(1.08) \\
\hline

\hline
\multirow{3}{*}{ASTGCN} & \multirow{2}{*}{\emph{Partial}} & \emph{Value} & 5.57(0.72) & 10.61(1.36) & & 6.14(1.29) & 8.95(2.35) & & 22.44(1.58) & 43.07(2.46) & & 3.60(0.60) & 6.05(1.13) \\
& & $\Delta_{partial}$ & 10.51 \% & 10.63 \% & & 35.24 \% & 38.11 \% & & 17.05 \% & 7.11 \% & & 3.74 \% & 3.77 \% \\
\cline{2-14}
& \emph{Oracle} & & 5.04(0.39) & 9.59(0.62)  & & 4.54(0.47) & 6.48(0.85) & & 19.17(0.91) & 40.21(2.02) &  & 3.47(0.50) & 5.83(0.99) \\
\hline

\hline
\multirow{3}{*}{MSTGCN} & \multirow{2}{*}{\emph{Partial}} & \emph{Value} & 4.78(0.43) & 9.35(0.75) & & 4.75(0.73) & 7.02(1.42) & & 18.96(1.21) & 40.13(2.67) & & 4.34(0.87) & 7.61(1.86) \\
& & $\Delta_{partial}$ & 6.45 \% & 4.70 \% & & 30.49 \% & 25.35 \% & & 8.90 \% & 6.05 \% & & 28.02 \% & 30.75 \% \\
\cline{2-14}
& \emph{Oracle} & & 4.49(0.31) & 8.93(0.50)  & & 3.64(0.41) & 5.60(0.82)  & & 17.41(0.74) & 37.84(1.88) &  & 3.39(0.52) & 5.82(1.06) \\
\hline

\hline
\multirow{3}{*}{LSTNet} & \multirow{2}{*}{\emph{Partial}} & \emph{Value} & 6.88(0.59) & 11.68(0.93)  & & 7.17(0.91) & 10.22(1.49)  & & 21.47(2.23)  & 42.62(3.32) &  & 4.97(0.72) & 8.27(1.43)  \\
& & $\Delta_{partial}$ & 19.65 \% & 18.94 \% & & 32.04 \% & 41.16 \% & & 13.71 \% & 15.12 \% & & 28.09 \% & 27.03 \% \\
\cline{2-14}
 & \emph{Oracle} & & 5.75(0.28) & 9.82(0.53)   & & 5.43(0.48) & 7.24(0.64)   & & 18.88(0.76)  & 37.02(2.47)  &  & 3.88(0.49) & 6.51(0.98) \\
\hline

\hline
\multirow{3}{*}{TCN} & \multirow{2}{*}{\emph{Partial}} & \emph{Value} & 6.41(0.61) & 11.34(0.86)  & & 6.82(0.62) &  9.91(1.25)  & & 23.31(1.57)  & 46.49(2.29) &  & 4.94(0.65) & 7.93(1.51)  \\
& & $\Delta_{partial}$ & 16.33 \% & 16.18 \% & & 45.10 \% & 44.88 \% & & 13.43 \% & 10.19 \% & & 33.15 \% & 30.21 \% \\
\cline{2-14}
& \emph{Oracle} & & 5.51(0.38) & 9.76(0.60)  & & 4.70(0.29) & 6.84(0.79)  & & 20.55(0.70)  & 42.19(1.98) &  & 3.71(0.56) & 6.09(1.13) \\
\hline

\end{tabular}

\end{table*}

\section{Performance Gap Analysis}
\label{sec:performance_gap}
In this section, we study the performance of an array of recent and popular MTSF   methods (MTGNN \cite{wu2020connecting}, ASTGCN \cite{Guo_Lin_Feng_Song_Wan_2019}, MSTGCN \cite{Guo_Lin_Feng_Song_Wan_2019}  LSTNet \cite{10.1145/3209978.3210006}, TCN \cite{DBLP:journals/corr/abs-1803-01271})  on the variable subset forecast problem. Then we further investigate the gaps that exist in current techniques that motivates our solution in Section \ref{sec:our_method}.

\subsection{Datasets} 
We consider 4 widely used datasets in the literature, described as follows - \textbf{(i) METR-LA} \cite{10.1145/2611567}: This dataset contains average traffic speed measured by 207 loop detectors on the highways of Los Angeles ranging from Mar 2012 to Jun 2012. \textbf{(ii) \footnote{https://www.nrel.gov/grid/solar-power-data.html}SOLAR}: This dataset contains the solar power output that was collected from 137 plants in Alabama State in 2007. \textbf{(iii) \footnote{https://pems.dot.ca.gov/}TRAFFIC}: This dataset contains road occupancy rates that were measured by 862 sensors in San Francisco Bay area during 2015 and 2016 . Since the default scale of a substantial fraction of values is of the order of $1e^{-3}$, we upscale (multiply the variable values) by $1e^{3}$. \textbf{(iv) \footnote{http://www.timeseriesclassification.com/description.php?Dataset=ECG5000}ECG5000}: This dataset from the UCR time-series Classification Archive consists of 140 electrocardiograms (ECG) with a length of 5000 each (we use it for the purpose of forecasting, as done by \cite{NEURIPS2020_cdf6581c}). Similar to \emph{TRAFFIC}, we upscale all variables by a factor of $10$.

    
    
    
Table \ref{table:stats_datasets} contains the important statistics for the datasets.

\subsection{Baseline Variants}
For each of the 5 forecast methods, we implement two baselines as described below. Both baselines are trained on the full training data but differ in terms of the inference procedure. 
    
    \textbf{Partial:} In this setting, we completely remove the data for the remaining $\mathcal{N} \setminus \mathcal{S}$ variables during test time forward pass. 
    This represents the performance of the forecast model on the VSF problem in the form it was originally proposed. 
    
    \textbf{Oracle: } Although infeasible in practice, this setting provides an upper bound to the results we seek to obtain and is solely for the purpose of comparison. During inference, we assume access to the data for all $\mathcal{N}$ variables, which allows the spatial module to make completely informed forward pass. 
    However the error metrics are computed only for the variables in $\mathcal{S}$, as desired for comparison. 

\subsection{Metrics:}
\label{sec:metrics} 
Representing the model forecast and the ground truth at time step $j$ with $\hat{Y}_{:,j}$ and $Y_{:,j}$ respectively, the error metrics can be defined as follows:

        \begin{equation}
        \label{eq:mae}
            \textsf{Mean Absolute Error (MAE)}  = \frac{1}{|\mathcal{S}|} \sum_{n=1}^{|\mathcal{S}|} \lvert \hat{Y}_{n,j} - Y_{n,j} \rvert
        \end{equation}

        \begin{equation}
        \label{eq:rmse} 
            \textsf{Root Mean Squared Error (RMSE)} = \sqrt{ \frac{1}{|\mathcal{S}|} \sum_{n=1}^{|\mathcal{S}|} ( \hat{Y}_{n,j} - Y_{n,j} )^2 }
        \end{equation}
    
    \textbf{Oracle Gap ($\Delta_{partial}$)} 
    We evaluate our proposed methodology by quantifying the performance relative to the \emph{Oracle} as follows:
        \begin{equation}
        \label{eq:delta}
            \Delta_{partial} = \frac{ E_{Partial} - E_{Oracle} }{ E_{Oracle} } \times 100
        \end{equation}
        Here \emph{E} is the placeholder 
        for the error metrics \emph{MAE} and \emph{RMSE}. For each technique $\Delta_{partial}$ represents the headroom that we aim to recover using the algorithms we propose in this paper.
            
For the error metrics, lower values are better. Also, note that in 
Equations \ref{eq:mae} and \ref{eq:rmse} the error is computed only over the variables that are present in $\mathcal{S}$ and {\em not} over all the variables $\mathcal{N}$.

\subsection{Experiment Setup}
During inference (ie testing), we consider $|\mathcal{S}|$ to be some percent $k$ of $|\mathcal{N}|$. We set $k =  15\%$ for all experiments unless stated otherwise (variation of $k$ is studied in section \ref{subsec:vary_subset_size}). 
For each trained model, we randomly sample the subset $\mathcal{S}$ $100$ times to perform the inference. This helps ensure a good coverage of  subsets of $\mathcal{N}$. 
To better estimate the uncertainty due to random initialization, we train $10$ models from scratch and report the mean and variance of the results . Therefore all the error metrics reported in our experiments involve $100 \times 10 = 1000$ runs over the test set. The error metrics are computed for forecast horizon length $Q$ at $12$, unless specified otherwise. 
 We use $70\%$ of the samples for training, $10\%$ for validation and $20\%$ for testing.  
Other hyperparameter details are provided in section \ref{subsec:hyperparams}.

\subsection{Performance of Current Forecast Models} 
The results are provided in table \ref{table:oracle_vs_subset_setting}. Note the substantial increment in the {\em Gap $\Delta_{partial}$ w.r.t Oracle}, sometimes even higher than $45\%$. Following the time series literature \cite{wu2020connecting, ijcai2019-264, NEURIPS2020_cdf6581c}, we note that this relative gap is large enough for this problem setup to be considered challenging and leaves room for large improvement over the Partial case. The high standard deviation in results is attributed to the variability in selecting $\mathcal{S}$. 
Similar to results reported in recent literature, we see Graph based methods perform better than the Deep Learning based methods in the {\em Oracle} case. In the {\em Oracle} setting, MTGNN has significantly better performance across all datasets except ECG5000 where its performance is comparable  to other techniques. Also, MTGNN sees the the largest drop in performance in the {\em Partial} setting as seen by the relatively large values of $\Delta_{partial}$. We hypothesize that MTGNN is able to model inter-variable relationships better that lead to it outperforming other techniques in the {\em Oracle} and the same reason causes its performance to degrade significantly when such information is not available in the {\em Partial} setting. For instance in the Traffic dataset, MTGNN oracle is almost $50\%$ better than other methods while its $\Delta_{partial}$ is the highest across all dataset, model combinations. We also observe that there is no one single model that has the lowest $\Delta_{partial}$ across all datasets.  This shows absence of data affects different models in complex ways. Hence a wrapper approach, like the one proposed in this paper (section \ref{sec:our_method}), can abstract  the complexity of the VSF problem to re-use SOTA MTSF algorithms in the VSF setting.  


\begin{figure}[t]
		\includegraphics[width=70mm,height=50mm]{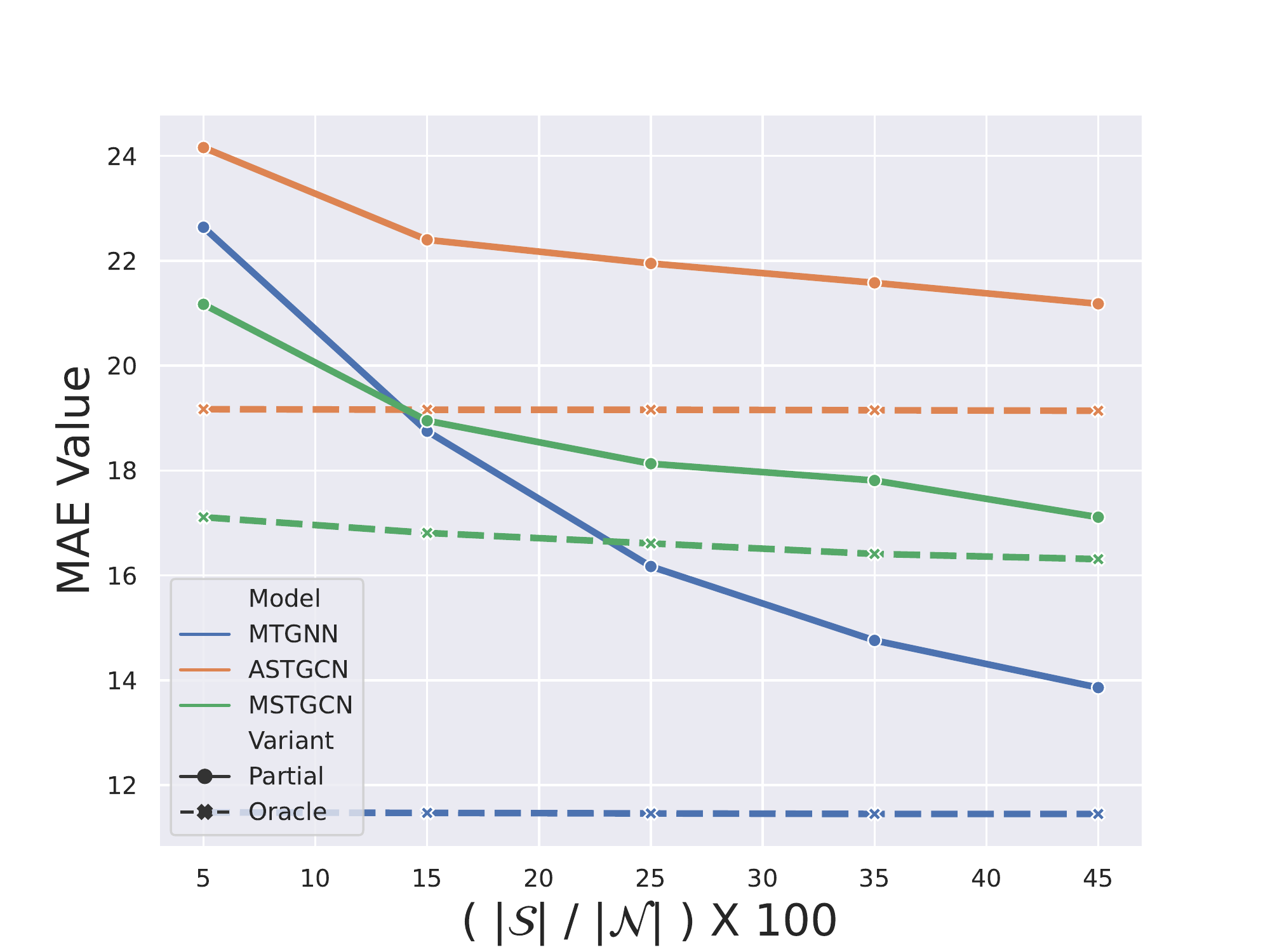}
		\label{subfig:varying_subset_size_mae}
	
	\caption{Variation of Error metrics with relative size of $\mathcal{S}$ for the Traffic dataset. The Gap closes with increasing size of $\mathcal{S}$. Solid lineplot is the \emph{Partial} setting, whereas the dashed correspond to \emph{Oracle} (nearly constant), for a given model.} 
	\label{fig:varying_subset_size_mae}
\end{figure}

\label{sec:key_observations} 
\subsubsection{On the Size of $\mathcal{S}$}
\label{subsec:vary_subset_size}

 In figure \ref{fig:varying_subset_size_mae} we show the MAE  on TRAFFIC dataset for \emph{Partial} and \emph{oracle} settings for various sizes of $\mathcal{S}$.
We see that the error decreases as the size of $\mathcal{S}$ increases since: (i) more data is available, (ii) with more variables in the input, the likelihood of co-occurence of two variables in $\mathcal{S}$ that have very high association also increases, thus allowing the spatial module to exploit the inter-dependencies in the input better and improve the overall forecast.\\

\subsubsection{On the apriori knowledge of $\mathcal{S}$ }
We conduct another oracle experiment where we assume apriori knowledge of  $\mathcal{S}$ and train the models on this data as well as further infer on the same. $50$ models with random selections of $\mathcal{S}$ are trained and the reported results are averaged. The remaining hyperparameters are kept same. Note that this setting is infeasible in practice and is solely for the purpose of analysis as we don't assume anything about $\mathcal{S}$ in this work. The results shown in table \ref{table:S_apriori} demonstrate that apriori knowledge of $\mathcal S$ does not necessarily help recover oracle performance. In the oracle case, the spatial module (which learns the inter-variable associations) has more information and can model the dynamics better during the forward pass, compared to this apriori setting. 
We also see that the $\mathcal{S}$ apriori performance is better than the {\em Partial} setting in Table\ref{table:oracle_vs_subset_setting} as the model parameters are trained to adjust to the given data.

\begin{table}[t]
\small
\caption{
Results for $\mathcal{S}$ apriori experiments (on ECG5000). Note: this is also an oracle setup and infeasible in practice, since we don't assume any knowledge about the subset $\mathcal{S}$ in this work. $\Delta_{\emph{$\mathcal{S}$ apriori}}$ represents the results for \emph{$\mathcal{S}$ apriori} setting and is computed in a similar manner as eq \ref{eq:delta}.} 
\label{table:S_apriori}

\centering
\setlength\tabcolsep{3.5pt}
\begin{tabular}{|c|c|cc|}

\hline

Model  & Setting & MAE  & RMSE  \\
\hline


\multirow{2}{*}{MSTGCN} & $\Delta_{\emph{$\mathcal{S}$ apriori}}$  & 5.38 \% & 4.89 \% \\ 
    \cline{2-4}
& \emph{Oracle} & 3.39(0.52) & 5.82(1.06) \\ 
\hline

\hline
\multirow{2}{*}{MTGNN} & $\Delta_{\emph{$\mathcal{S}$ apriori}}$ &  4.30 \% & 5.27 \% \\
    \cline{2-4}
& \emph{Oracle} & 3.43(0.54) & 5.94(1.08) \\ 
\hline

\end{tabular}

\end{table}

\subsubsection{On the importance of inter-variable associations}
\label{subsec:identity_adjacency}
\noindent
Since $\mathcal{S}$ is selected at random, a natural question to ask is - "what if we remove the spatial module and let the models learn solely based on temporal patterns?". We design an experimental setting for the graph based forecast models (shown to be superior in performance), where the adjacency matrix during training as well as inference is replaced with an Identity matrix $I_{\mathcal{N} \times \mathcal{N}}$. The results shown in table \ref{table:identity_adjacency} verify the importance of the associations, either learnt or predefined, and consequently the flow of information via message passing.


\section{Proposed Solution}
\label{sec:our_method}
In the previous section, we showed that current MTSF algorithms underperform in the VSF formulation primarily because of absence of data of other $\mathcal{N -S}$ variables and the complex relationships of variables in $\mathcal{S}$ to the ones in  $\mathcal{N -S}$. In this section, we build on these observations to propose a simple, yet effective, wrapper technique to address the Variable Subset Forecast problem.

\subsection{kNN Retrieval over Variable Subset $\mathcal{S}$}
\label{subsec:borrowing_data}
The extensive evaluations in the previous section have showed that the remaining variables play an important role in the process. We thus resort to a \emph{retrieval} based mechanism to first borrow the data for the remaining $\mathcal{N} - \mathcal{S}$ variables for the given test instance and further perform the forward pass through $f$. The underlying idea is that - with the original test instance data of $\mathcal{S}$ and the borrowed data for $\mathcal{N} - \mathcal{S}$, the spatial module and thereby the entire forecast model $f$ can make much more informed decision. 
Recently, similar non parametric k-NN based retrieval techniques have also been used to significantly improve Neural Machine Translation, Question Answering and Language Modeling\cite{khandelwal2021nearest,DBLP:journals/corr/abs-2010-11856,bapna-firat-2019-non, xu-etal-2020-boosting} in NLP.

Formally defining, given a test instance $X^{test}$  which can be represented as a $3D$ tensor of shape $P \times |\mathcal{S}| \times D$ ($P$ being time steps, $|\mathcal{S}|$ being size of selected subset and $D$ being the dimension of each node), we compute its distance to the instances in entire training data, to obtain the neighbors. The choice of this distance is user-dependent and there are many different options available in the literature such as \textit{multivariate dynamic time warping} \cite{time_series_hashing}, \textit{global alignment kernel} \cite{10.5555/3104482.3104599} etc., however a simple \textit{euclidean distance} \cite{time_series_hashing} type function has been shown to work well and has an added benefit of being substantially faster than others. Simply and more intuitively, we can understand it as operating over the distance of the flattened $3D$ tensor into a vector of length $P * |\mathcal{S}| * D$ (where $*$ is the multiplication over reals). 

The distance between two time series instances $X^{\prime}$ and $X^{\prime\prime}$ can be computed as:
\begin{equation}
\label{eq:dist}
    D(X^{\prime}, X^{\prime\prime}) = \frac{1}{P * |\mathcal{S}| * D} \sum_{p=1}^{P} \sum_{s=1}^{|\mathcal{S}|} \sum_{d=1}^{D} \lvert X^{\prime}_{p,s,d} - X^{\prime\prime}_{p,s,d} \rvert ^b    
\end{equation}
where $b$ is the exponent factor. We also observed in our ablation experiments that $b = 0.5$ works best.
Denoting the actual training set as $\mathcal{T}$, the nearest neighbor to $X^{test}$ is computed as:
\begin{equation}
\label{eq:neighbor}
  X^{NN} = \argmin_{X^{\prime} \in \mathcal{T}} D(X^{test}, X^{\prime}_{[:, \mathcal{S}, :]})
\end{equation}
where $[:, \mathcal{S}, :]$ is standard data indexing from tensors. 
Lastly, we construct a new instance $X^{new}$, such that $X^{new}_{[:, \mathcal{S}, :]} = X^{test}$ and $X^{new}_{[:, \mathcal{N} \setminus \mathcal{S}, :]} = X^{NN}_{[:, \mathcal{N} \setminus \mathcal{S}, :]}$. The output obtained from the forecast model $f$ over $X^{new}$ in the dimensions corresponding to $\mathcal{S}$ is subsequently used to compute the error metrics. 


\begin{table}[t]
\small

\caption{Results on MTGNN model for Section \ref{subsec:identity_adjacency} (Identity Matrix Case). $\Delta_{Identity}$ represents the results for \emph{Identity Matrix} setting and is computed in a similar manner as eq \ref{eq:delta}.} 
\label{table:identity_adjacency}

\centering
\setlength\tabcolsep{3.5pt}
\begin{tabular}{|c|c|cc|}

\hline
Dataset & Setting &  MAE  & RMSE \\

\hline
\multirow{3}{*}{METR-LA} & $\Delta_{Identity}$ & 47.27 \% & 35.92 \% \\ 
 & $\Delta_{partial}$ & 30.08 \% & 23.43 \%  \\ 
 \cline{2-4}
 & \emph{Oracle}  & 3.49(0.25) & 7.21(0.50) \\ 
\hline

\hline
\multirow{3}{*}{SOLAR} & $\Delta_{Identity}$ & 104.76 \% & 64.59 \% \\ 
 & $\Delta_{partial}$ & 44.89 \% & 29.61 \% \\ 
 \cline{2-4}
 & \emph{Oracle}  &  2.94(0.27) & 4.66(0.57) \\
\hline

\hline
\multirow{3}{*}{TRAFFIC} & $\Delta_{Identity}$ & 155.54 \% & 83.87 \% \\ 
 & $\Delta_{partial}$ & 62.18 \% & 39.95 \%  \\ 
 \cline{2-4}
 & \emph{Oracle}  &  11.45(0.57) & 27.48(2.14) \\ 
\hline

\end{tabular}

\end{table}

\subsection{Weighted Ensembling}
We further explore the usage of an ensemble over multiple neighbors to improve performance. The method to retrieve neighbors remains same as in equation \ref{eq:neighbor}, with the only difference that top-$m$ neighbors with least distance are selected. Correspondingly, $X^{new_i}$, $i \in \{1, 2, ...m\}$, are created where in each of these, the data for the $\mathcal{S}$ variables is same as given test instance $X^{test}$ and the data for remaining variables is replaced with the $i^{th}$ neighbor $X^{NN_i}$, one by one. 

\subsubsection{Uniform Weighting (UW):} The most standard way of assigning weights to the $m$ forecasts is by averaging.

However, out of these $m$ forecasts obtained using data of $m$ neighbors, it is evident that some outputs will be closer to the actual forecast for $X^{test}$, call it $Y^{test}$, compared to others, indicating a weighted average of forecasts would work better. One can utilize simple mathematical functions to complex neural networks in order to predict these scalar weights. 

\subsubsection{Direct Distance Weighting (DDW):} 
A very standard way to assign the weights to these neighbors is by simply computing softmax of the negative of the distance computed in equation \ref{eq:dist}, as follows:
\begin{equation}
\label{eq:ensemble_weights_ddw}
     w_i = \frac{e^{-D(X^{test}, X^{NN_i}_{[:, \mathcal{S}, :]}) / \tau }}{ \sum_{j=1}^{m} e^{-D(X^{test}, X^{NN_j}_{[:, \mathcal{S}, :]}) / \tau } }     
\end{equation}
where $\tau$ is the \emph{temperature} parameter and $w_i$ is the weight given to the forecast of the $i^{th}$ neighbor. However, this has multiple caveats. Firstly, this scheme does not take into account the spatial module that governs the interactions amongst the variables. Secondly, it also noteworthy that if two instances are close based on the distance of few variables does not necessarily mean the instances will be close when all $\mathcal{N}$ variables are considered, because of the bias in retrieval. 

\subsubsection{Forecast Distance Weighting (FDW) (Algorithm \ref{algo:our_method}):}
 We retrieve the top-$m$ neighbors using equations \ref{eq:dist} and \ref{eq:neighbor}.
 However, to compute better weights, we propose to use a parametric and complex neural network, for which $f$ seems to be a reasonable choice. Instead of computing the vicinity of training instances to $X^{test}$ (in some other space), we measure the discrepancy of the forecast outputs between $X^{new_i}$ and the corresponding $X^{NN_i}$, $\forall i \in \{1, 2 ... m\}$. Note that $X^{new_i}$ is a combination of data from $X^{test}$ and $X^{NN_i}$. Making a forward pass through $f$ for both of these inputs will provide a forecasts that can be represented as matrices of shape $|\mathcal{N}| \times Q$, $Q$ being forecast output length. These forecasts are obtained as a result of the combination of complex operations that capture global or pairwise interactions in conjunction with RNNs (or temporal CNNs). The forward passes check the compatibility of the data borrowed for the remaining $\mathcal{N} \setminus \mathcal{S}$ variables to the original data of $\mathcal{S}$ variables in $X^{test}$. If both the forecasts are close (via some function), then we assume that the corresponding neighbor's borrowed data is a good proxy for missing data in $X^{test}$. The entire retrieval and ensemble weighting procedure is   formulated as follows.

First, lets denote the forecasts of $X^{new_i}$ and $X^{NN_i}$, ie $f(X^{new_i})$ and $f(X^{NN_i})$, as $Y^{new_i}$ and $Y^{NN_i}$ respectively. The distance function we use is akin to the mean absolute error metric, as shown below:
\begin{equation}
\label{eq:ensemble_weights_fdw}
    D_F(Y^{new_i},Y^{NN_i}) = \frac{1}{Q * |\mathcal{S}|} \sum_{q=1}^{Q} \sum_{s=1}^{|\mathcal{S}|} \Bigg\lvert \frac{Y^{new_i}_{q,s} - Y^{NN_i}_{q,s}}{q}  \Bigg\rvert
\end{equation}
where $D_F$ simply denotes the distance function for forecasts. Note the use of  $\mathcal{S}$ as we seek to obtain the discrepancy in the corresponding dimensions. The factor $q$ in denominator ensures that values in longer horizons don't dominate the computation and serves as a normalizing factor (supporting experiments are provided in appendix section \ref{subsec:gap_subset_oracle_across_horizons} ).
Finally, the weights are computed as: 
\begin{equation}
\label{eq:ensemble_weights_forecast_dis}
     w_i = \frac{e^{ -D_F(Y^{new_i},Y^{NN_i}) / \tau }}{ \sum_{j=1}^{m} e^{ -D_F(Y^{new_j},Y^{NN_j}) / \tau } }     
\end{equation}

\begin{algorithm}[t]
\caption{\textsc{Proposed Forecast Distance Weighting (FDW)}}
\label{algo:our_method}
\begin{algorithmic}[1]
    
    \REQUIRE Forecast model $f$, Training set $\mathcal{T}$, Test Instance $X^{Test}$, Forecast Distance function $D_F$, \# Neighbors $m$, exponent $b$, temperature $\tau$
    
    \ENSURE Ensembled Forecast $\hat{Y}^{new}$ for test instance $X^{Test}$
    %
    \STATE \textsl{\# Obtain Set of Nearest Neighbors using Eq \ref{eq:dist} and \ref{eq:neighbor} }
    \STATE $\mathcal{K}^{NN} = \{X^{NN_i} | i \in [\![1...m]\!] , X^{Test}, \mathcal{T}, b \}$
    \STATE
    \STATE $\mathcal{D} = []$ \textsl{\# List of Forecast Distances Eq \ref{eq:ensemble_weights_fdw} }
    \STATE $\mathcal{Y} = []$ \textsl{\# List of Forecasts using $m$ neighbors}
    \STATE
    \FOR{$X^{NN_i} \in \mathcal{K}^{NN}$}  
    \STATE $X^{new_i} = X^{test} || X^{NN_i}$  \textsl{\# Create new instance, section \ref{subsec:borrowing_data}}
    \STATE $Y^{new_i}, Y^{NN_i} = f(X^{new_i}), f(X^{NN_i})$    
    \STATE $\mathcal{Y}.append(Y^{new_i})$
    \STATE $\mathcal{D}.append(D_F(Y^{new_i}, Y^{NN_i}))$ \textsl{\# Forecast Distance (Eq \ref{eq:ensemble_weights_fdw})}
    \ENDFOR
    \STATE
    \STATE $\mathcal{W} = \{w_i | i \in [\![1...m]\!] , \mathcal{D}, \tau \}$ \textsl{\# FDW (Eq \ref{eq:ensemble_weights_forecast_dis}) }
    \STATE 
    \STATE $\hat{Y}^{new} = 0$
    \FOR{$i \in [\![1...m]\!]$}
    \STATE $\hat{Y}^{new} \mathrel{+}= \mathcal{W}[i] \mathcal{Y}[i]$
    \ENDFOR
    \STATE Return $\hat{Y}^{new}_{[:, \mathcal{S}]}$
\end{algorithmic}
\end{algorithm}

\section{Empirical Evaluation}
\label{sec:empirical_evaluation}
We use $\Delta_{Ensemble}$ to quantify the performance of the proposed FDW ensemble method relative to the \emph{Oracle}, defined as follows:

\begin{equation}
\label{eq:delta_ensemble}
    \Delta_{Ensemble} = \frac{ E_{Ensemble} - E_{Oracle} }{ E_{Oracle} } \times 100
\end{equation}
Here, $E_{Ensemble}$ are the errors for the proposed ensemble case and $E_{Oracle}$ is retrieved from Table \ref{table:oracle_vs_subset_setting}. The results are provided in Table \ref{table:ewp_main}. Details regarding the hyperparameters in this approach are provided in section \ref{subsec:hyperparams}. We see that across all methods and datasets, the \emph{FDW} ensemble substantially recovers the performance drop we see in table \ref{table:oracle_vs_subset_setting}. For a majority of the cases, we see the relative gap with the oracle is less than $5\%$.  
 

\begin{table*}[tbp]

\caption{Relative performance, $\Delta_{Ensemble}$, of our proposed FDW ensemble method w.r.t \emph{Oracle}. We substantially improve over the $\Delta_{partial}$ 
in Table \ref{table:oracle_vs_subset_setting}.} 
\label{table:ewp_main}

\centering
\setlength\tabcolsep{3.5pt}
\begin{tabular}{|l|ccc|ccc|ccc|cc|}

\hline
  \multirow{2}{0cm}{Forecast  Model} & \multicolumn{3}{c}{\bf \emph{METR-LA}}  &  \multicolumn{3}{c}{\bf \emph{SOLAR}} & \multicolumn{3}{c}{\bf \emph{Traffic}} & \multicolumn{2}{c|}{\bf \emph{ECG5000}} \\
\cline{2-4}   \cline{5-7} \cline{8-10} \cline{11-12}
& MAE  & RMSE  &  &  MAE  & RMSE &  &  MAE  & RMSE & & MAE & RMSE \\

\hline
\textbf{MTGNN} & 7.50 \% & 5.46 \%  & & 30.05 \% & 25.45 \%  & & 1.13 \% & 2.01 \% & & 2.14 \% & 3.11  \% \\
\hline


\hline
\textbf{ASTGCN} & 1.88 \% & 2.79 \%  & & 4.32 \% & 4.21 \%  & & 0.17 \% & 0.42 \% & & 0.26 \% & 1.03 \% \\
\hline


\hline
\textbf{MSTGCN} & 0.71 \% & 0.56 \%  & & 9.50 \% & 7.65 \%  & & 0.78 \% & 1.01 \% & & 1.13 \% & 0.59 \% \\
\hline


\hline
\textbf{LSTNet} & 1.39 \% & 1.02 \%  & & 11.78 \% & 14.22 \%  & & 0.47 \% & 0.54 \% & & 1.36 \% & 0.73 \% \\
\hline


\hline
\textbf{TCN} & 1.98 \% & 1.53 \%  & & 19.57 \% & 18.12 \%  & & 0.38 \% & 0.21 \% & & 1.07 \% & 0.31 \% \\
\hline


\end{tabular}

\end{table*}



\begin{figure}[t]

	\includegraphics[scale = 0.25]{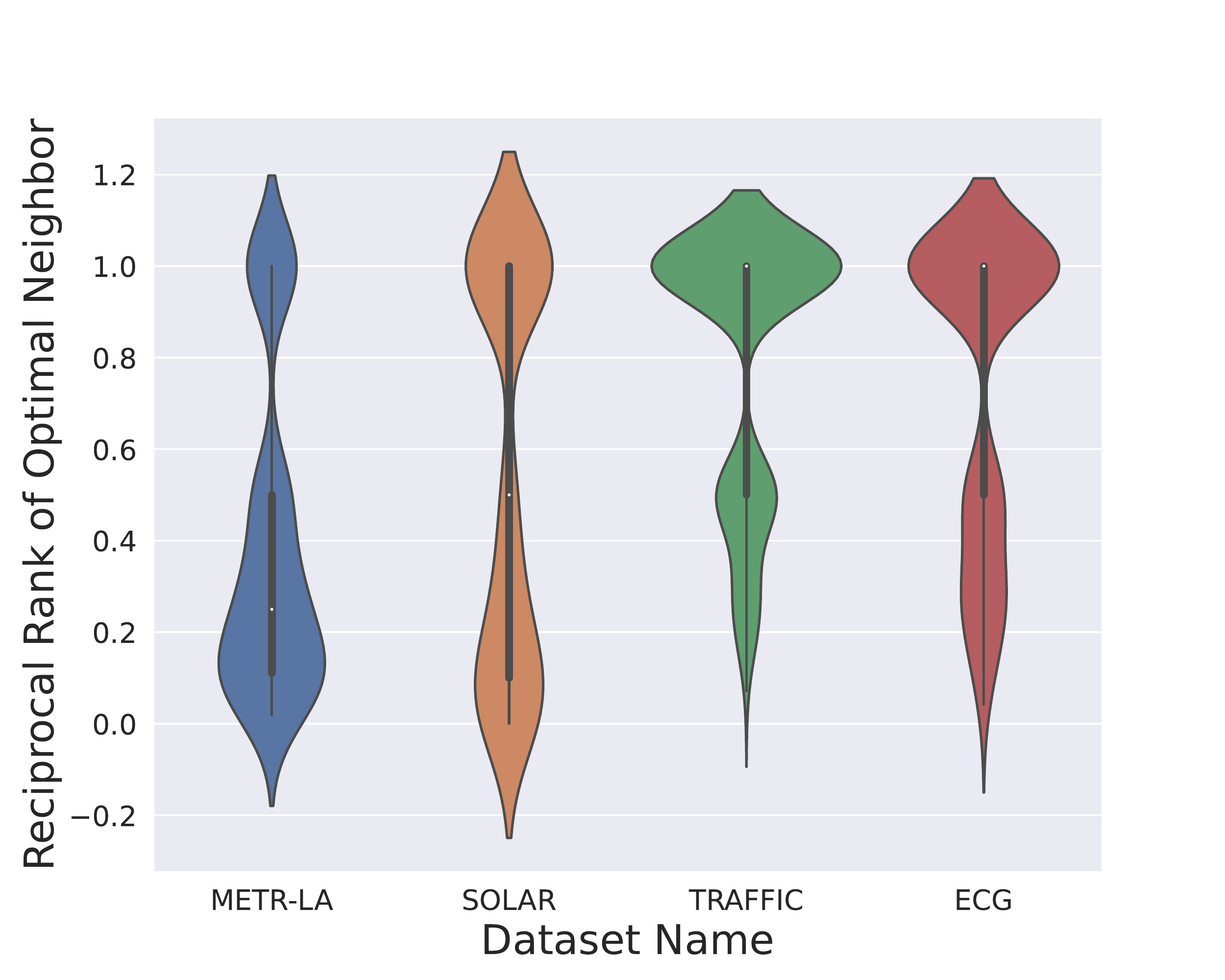}
		 	\caption{ Reciprocal Rank of Optimal Nearest Neighbor}
		\label{fig:analysis_1_optimal_NN}
\end{figure}

 We also observe that the performance of all models on Solar is relatively poor compared to that of Traffic (and others). We analyse the datasets using a ranking measure called {\em reciprocal rank} to uncover the reasons of this performance parity as follows.  
For every validation instance, first we find the optimal (nearest) neighbor by using all $\mathcal{N}$ variables. We then find the rank of this optimal neighbor in the list obtained by retrieving neighbors using the distance on the $\mathcal{S}$ variables as in Equation~\ref{eq:dist}. Figure~\ref{fig:analysis_1_optimal_NN} shows the reciprocal of rank of the optimal neighbor for the four datasets. Larger values of reciprocal rank (close to 1) is better and means the optimal neighbor can be found relatively close to the corresponding validation instance.  Small (close to zero) reciprocal ranks show that the optimal neighbor is far away from the corresponding validation instance.  For  \emph{SOLAR} and \emph{METR-LA} datasets, the one mode of the distribution substantially lies close to zero. \emph{Traffic} and \emph{ECG} datasets, on the other hand, have a much better reciprocal rank distribution.


\begin{table}[h]
\small

\caption{Hyperparameter Ablation Study of \emph{FDW} 
 for MSTGCN on ECG5000 and MTGNN on METR-LA.} 
\label{table:ablation_our_method}

\centering
\setlength\tabcolsep{3.5pt}
\begin{tabular}{|l|rr|rr|}

\hline
   &  \multicolumn{2}{c|}{\bf \emph{ECG5000}}  &  \multicolumn{2}{c|}{\bf \emph{METR-LA}}  \\
\cline{2-5}  
 &  MAE  & RMSE & MAE & RMSE \\ 
\hline

\emph{Main Results ($b=0.5$, $\tau=0.1$)} & 1.13 \% & 0.59 \% & 7.50 \% & 5.46 \% \\
\hline 

\multicolumn{5}{|c|}{\bf \emph{Hyperparameters}} \\
\hline
\emph{Exponent $b=0.33$} & 0.54 \% & 0.31 \% & 7.57 \% & 5.59 \% \\
\emph{Exponent $b=1$} & 1.16 \% & 0.97 \% & 10.66 \% & 10.24 \% \\
\emph{Exponent $b=2$} & 1.27 \% & 1.04 \% & 21.65 \% & 18.76 \% \\
\hline
\emph{Temperature $\tau=0.01$} & 1.32 \% & 1.26 \% & 6.98 \% & 6.32 \% \\
\emph{Temperature $\tau=1$} & 2.44 \% & 2.43 \% & 7.67 \% & 5.75 \% \\
\emph{Temperature $\tau=10$} & 0.66 \% & 0.58 \% & 7.55 \% & 5.77 \% \\
\hline

\end{tabular}

\end{table}
\subsection{Retrieval and Weighting Hyperparameters}
In this section, we perform an ablation study on two important hyperparameters  (i) exponent $b$ used in equation \ref{eq:dist} and temperature $\tau$ in \emph{FDW}. The experiments are performed for \textit{MSTGCN} model on \textit{ECG5000} and \textit{MTGNN} model on \textit{METR-LA}. The values of $\Delta_{Ensemble}$ are reported in Table \ref{table:ablation_our_method} where \emph{Main Results} denote the actual results borrowed from Table \ref{table:ewp_main}. While accounting for a specific hyperparameter, the rest are kept same. For example, if exponent $b$ is varied, then temperature $\tau$ is kept to it best value. The optimal hyperparameters were selected by search over the validation set reported in Appendix \ref{subsec:hyperparams}. 
First we see that as we increase $b$ performs drops rapidly. This can be explained by the construction of the distance function in Equation~\ref{eq:dist}. With large values of $b$ the elements in the summation that are large will contribute disproportionately and hence the distance would be dominated by a few dimensions where the gaps are large.   We however do not see a specific  pattern emerge for the temperature $\tau$ parameter.

\subsection{Comparison to k-NN method of \cite{tajmouati2021applying} }
Recent work by \cite{tajmouati2021applying} used k-NN algorithm to find nearest neighbors 
for a test instance and directly use the corresponding neighbors' forecast as the test time forecast, weighted via \emph{Rank Order Centroid}. Other works such as \cite{10.1145/3194452.3194467} also leverage similar strategies for financial time series data. 
We thus compare our approach to more recent \cite{tajmouati2021applying} (henceforth referred as \emph{Baseline}). 
Results in Table \ref{table:comparison_baseline_kNN} clearly demonstrates the superiority of performing forward pass through the forecast models so that spatial modules can properly exploit the underlying associations in order to generate informed outputs.

\begin{table}[h]
\scriptsize

\caption{Comparison of \emph{FDW} to Baseline \cite{tajmouati2021applying}. Each cell contains MAE / RMSE values of $\Delta_{Ensemble}$ (in \%) .} 
\label{table:comparison_baseline_kNN}

\centering
\begin{tabular}{|l|c|c|c|c|c|}

\hline
\textbf{ECG5000} &  MTGNN  & ASTGCN & MSTGCN & LSTNet & TCN \\
\hline
\emph{\emph{FDW}} & \textbf{2.14 / 3.11} & \textbf{0.26 / 1.03} & \textbf{1.13 / 0.59} & \textbf{1.36 / 0.73} & \textbf{1.07 / 0.31} \\
\hline
\emph{Baseline\cite{tajmouati2021applying}}  & 16.54 / 11.02 & 15.34 / 13.36 & 18.39 / 13.55 & 3.30 / 1.40 & 8.03 / 8.38 \\
\hline

\hline
\textbf{METR-LA} &  MTGNN  & ASTGCN & MSTGCN & LSTNet & TCN \\
\hline
\emph{\emph{FDW}} & \textbf{ 7.50 / 5.46 } & \textbf{ 1.88 / 2.79 } & \textbf{ 0.71 / 0.56 } & \textbf{ 1.39 / 1.02 } & \textbf{ 1.98 / 1.53 }  \\
\hline
\emph{Baseline\cite{tajmouati2021applying} }  & 70.23 / 59.27 & 18.02 / 19.61 & 32.70 / 28.72 & 3.51 / 16.99 & 8.02 / 17.71 \\
\hline


\end{tabular}

\end{table}

\section{Retrieval Ablations} 
\label{subsec:retrieval_ablations}
\subsection{Number of neighbors}
\label{subsec:vary_num_neighbors}
\begin{figure}[h]
	
	
	\begin{subfigure}[l]{0.23\textwidth}
		\raggedleft
		\includegraphics[width=45mm,height=25mm]{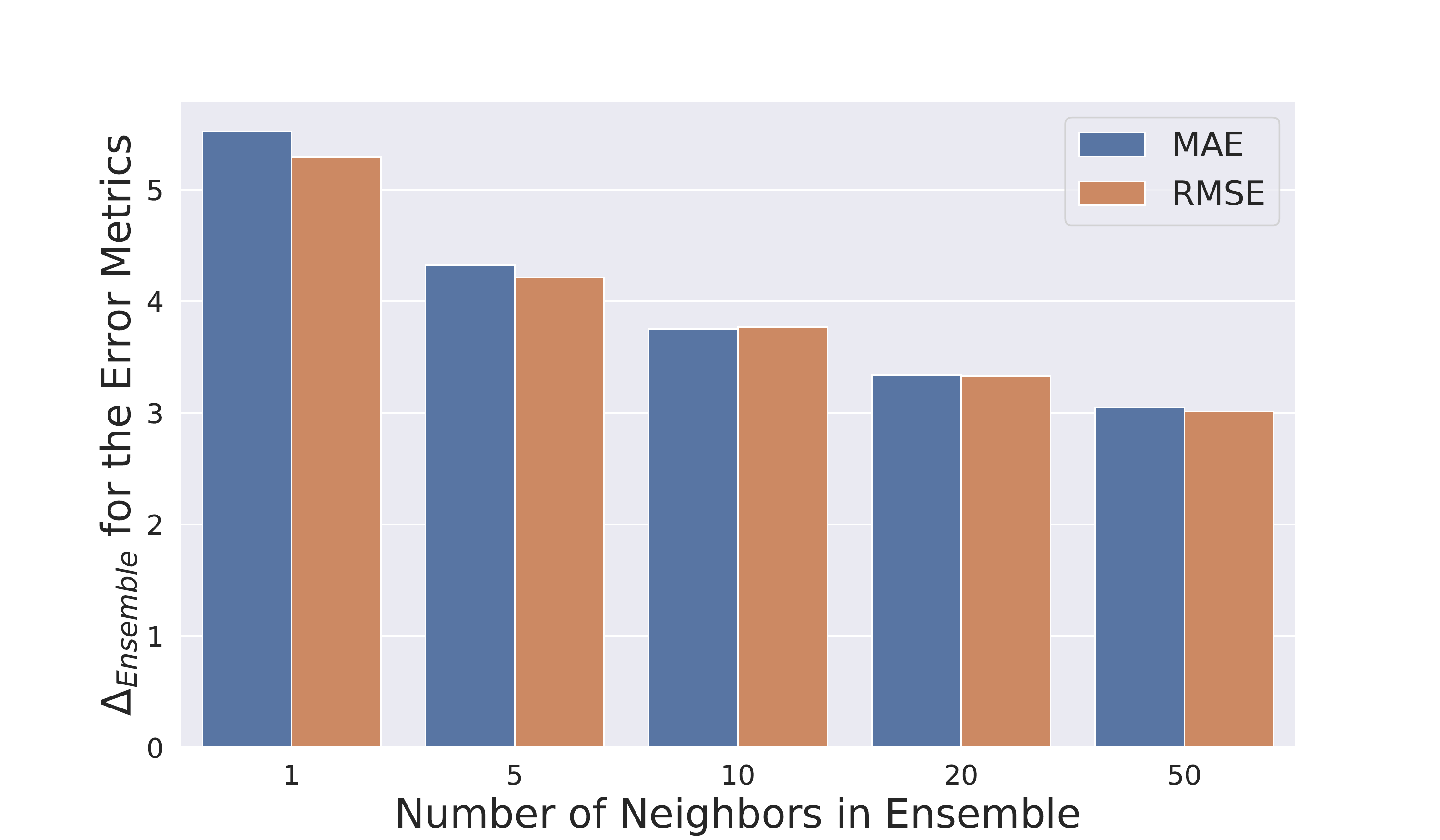}
		\caption{Vary Number of Neighbors}
		\label{subfig:vary_neighbors}
	\end{subfigure}%
	\begin{subfigure}[c]{0.23\textwidth}
		\includegraphics[width=45mm,height=25mm]{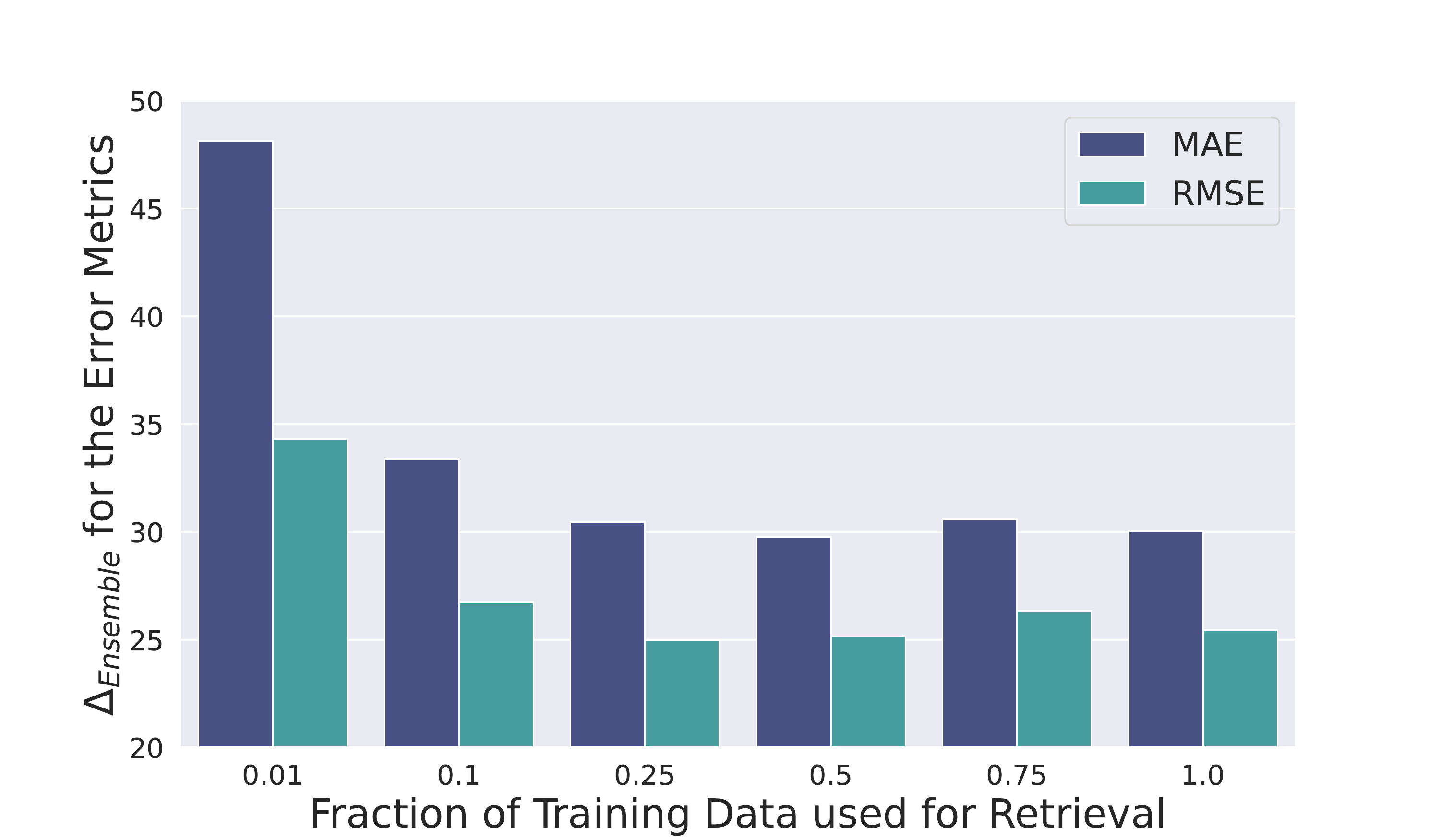}
		\caption{Vary Retrieval Set Size}
		\label{subfig:vary_retrieval_set}
	\end{subfigure}%
	
	\caption{Variation of $\Delta_{Ensemble}$ with number of neighbors (left subfigure) and size of \emph{Retrieval Set} w.r.t Training Set (right subfigure). 
	}	
	\label{fig:retrieval_analysis}
\end{figure}

We analyse the variation of $\Delta_{Ensemble}$ with changing number of neighbors in figure \ref{subfig:vary_neighbors} (for \textit{ASTGCN} on \textit{SOLAR}).
Note the \textit{strictly decreasing} bar sizes. This is due to the weight computation mechanism in \emph{FDW}. 
As we increase the number of neighbors, we also increase the likelihood of obtaining a neighbor which has very low discrepancy in the forecast w.r.t $X^{test}$ (eq \ref{eq:ensemble_weights_fdw}) and thus improve the overall ensemble forecast result whereas a large fraction of the obtained neighbors 
are assigned almost $0$ weight (the mechanism in eq \ref{eq:ensemble_weights_forecast_dis}). Furthermore, we have diminishing reductions in $\Delta_{Ensemble}$ with increasing number of neighbors. 
For our main experiments, we use $5$ neighbors as it provides a reasonable runtime  and performance tradeoff.

\subsection{Varying size of Retrieval Set}
To study the effect of the size of the source set over which retrieval is performed, we perform an experiment to observe  $\Delta_{Ensemble}$ for the error metrics against the fraction of  data used in the retrieval step. Figure \ref{subfig:vary_retrieval_set} (MTGNN over SOLAR) shows that this follows almost an \emph{asymptotic} behaviour, where the $\Delta_{Ensemble}$ improves sharply in smaller fractions and then improves gradually with increasing size of retrieval set. At \emph{quarter} size of the data, we see that the performance is very close to complete training set. Due to the complexity of the model, dataset combinations, we note that the fraction at which original performance is recovered can vary.

\begin{table}[t]
\small
\caption{Comparison of different Ensemble techniques by stating the \emph{MRR} (as defined in section \ref{subsec:compare_ensembles}) and $\Delta_{Ensemble}$ (put as $\Delta$ for brevity) RMSE values. The absence of MRR in \emph{UW} is because all neighbors are given equal weight, thus sorting the neighbor list isn't feasible due to floating point precision issues.
} 
\label{table:mrr_eq_simple_and_ours}

\centering
\begin{tabular}{|l|c|c|c|c|c|c|}

\hline
 & &  MTGNN  & ASTGCN & MSTGCN & LSTNet & TCN \\
\hline
\multirow{2}{*}{\emph{UW}} & MRR & - & - & - & - & - \\
& $\Delta$ & 2.83 \% & \textbf{0.31} \% & 1.24 \% & 0.94 \% & 0.82 \% \\
\hline
\multirow{2}{*}{\emph{DDW}} & MRR & 0.4532 & 0.4358 & 0.4439 & 0.4593 & 0.4560 \\
& $\Delta$ & 2.81 \% & 0.45 \% & 1.53 \% & 1.12 \% & 0.79 \% \\
\hline
\multirow{2}{*}{\emph{FDW}} & MRR & \textbf{0.4685} & \textbf{0.4591} & \textbf{0.4512} & \textbf{0.4729} & \textbf{0.4677} \\
 & $\Delta$ & \textbf{2.01} \% & 0.42 \% & \textbf{1.01} \% & \textbf{0.54} \% & \textbf{0.21} \% \\
\hline

\end{tabular}

\end{table}

\section{Weighting Ablations} 
\label{subsec:compare_ensembles}
In this section, we compare the performance of the three weighting methods : UW, DDW, FDW through both an intrinisic metric (mean reciprocal rank) and extrinsic metric (RMSE). From the list of $m$ retrieved neighbors, we first find the ideal neighbor as the one that produces the least error with respect to the ground truth. Therefore the ideal neighbor is model dependent.  We induce rankings of DDW (using Equation~\ref{eq:ensemble_weights_ddw}) and FDW (using Equation~\ref{eq:ensemble_weights_forecast_dis}). The results provided in Table \ref{table:mrr_eq_simple_and_ours}, over the validation set of \emph{Traffic}, demonstrate that the ideal neighbor in the retrieved set of size $m$ is ranked higher for the proposed eq \ref{eq:ensemble_weights_forecast_dis}. Correspondingly we see that the $\Delta_{Ensemble}$ for RMSE values of FDW is much better compared to DDW and that compared to UW.  This shows using the $f$ to compute the weights of the ensemble helps combine predictions more effectively. Table~\ref{table:comparison_eq_simple_and_ours} reports the MAE and RMSE values for the three weighing schemes aross 2 more combinations of dataset, models. The trend is consistent that FDW is able to significantly outperform the other methods.


\begin{table}[t]
\small
\caption{Comparison of different Ensemble Weighing techniques via corresponding $\Delta_{Ensemble}$ for each.}
\label{table:comparison_eq_simple_and_ours}

\centering
\begin{tabular}{|l|cc|cc|}
\hline
 & \multicolumn{2}{c|}{\emph{MSTGCN (ECG5000)}} & \multicolumn{2}{c|}{\emph{MTGNN (METR-LA)}}\\
 \cline{2-5}  
 &  MAE  & RMSE & MAE & RMSE \\ 
\hline
\emph{\emph{UW}} & 2.21 \% & 1.92 \% & 8.17 \% & 5.72 \% \\
\emph{\emph{DDW}} & 2.67 \% & 2.78 \% & 7.71 \% & 5.76 \% \\
\emph{\emph{FDW}} & \textbf{1.13} \% & \textbf{0.59} \% & \textbf{7.50} \% & \textbf{5.46} \% \\
\hline
\end{tabular}

\end{table}

\begin{table}[h]
\small

\caption{Comparison of our proposed technique \emph{FDW} to Popular Data Imputation Techniques - kNNE and IIM. The numbers in the table represent the values of $\Delta_{Ensemble}$ for both MAE/RMSE in \%.} 
\label{table:baseline_imputation}

\centering
\begin{tabular}{|l|c|c|c|c|}

\hline
\cline{1-5}  
 &  METR-LA & SOLAR & TRAFFIC & ECG5000 \\
\hline
\emph{FDW} & \textbf{7.50} / \textbf{5.46} & \underline{30.05} / \underline{25.45} & \underline{1.13} / \textbf{2.01} & \textbf{2.14} / 3.11 \\
\hline
\emph{kNNE} & \underline{8.36} / \underline{9.80} & 31.66 / 27.40 & \textbf{1.01} / \underline{2.42} & \underline{2.24} / \underline{3.05} \\
\emph{IIM}  & 9.55 / 10.85 & \textbf{30.04} / \textbf{24.94} & 3.39 / 3.83 & 2.68 / \textbf{3.04} \\
\hline


\end{tabular}

\end{table}

\section{Comparison to Data Imputation Techniques}
\label{sec:comparison_imputation}
As argued previously in section \ref{sec:intro}, a large literature of data imputation baselines is not applicable because of the heavy reliance on the temporal locality of the time series of each variable, which is absent altogether in our problem statement. 
For completeness, we provide a comparison to two popular techniques that can be applied in this setup without significant modifications: \textit{kNNE} \cite{knne_imputation} and IIM \cite{zhang2019learning}. The results on MTGNN model across the datasets are provided in table \ref{table:baseline_imputation} with the number of retrieved neighbors kept to $5$ for all methods in order to provide a fair comparison. Firstly, note that \emph{FDW} consistently outperforms or nearly matches the best case performance in any column of table \ref{table:baseline_imputation}. Furthermore, the imputation baselines are inconsistent in performance, for eg - \emph{IIM} performs slightly better in RMSE of SOLAR, but is worse on METR-LA and other two datasets. Similarly, \emph{kNNE} is close in performance on TRAFFIC, but is outperformed by our method \emph{FDW} on METR-LA nd SOLAR, which are the more difficult datasets compared to ECG5000 and TRAFFIC. One of the primary reasons that \emph{FDW} usually outperforms the imputation baselines is the novel solution to the biased retrieval problem.


\section{Correlated Failures}
\label{subsec:controlled_S_selection}
In the previous sections, we assumed that variables are absent at random during inference time. However, in real world settings, multiple sensor failures are not random but often correlated.  For instance, sensor values could be correlated and failures also correlated if they are spatially close. In this experiment, we simulate this condition by
 by first constructing $C$ disjoint clusters of $\mathcal{N}$. To generate these clusters, we apply \emph{DBSCAN} algorithm over the pairwise distance matrix, $\mathcal{D}$, where $\mathcal{D}_{ij} = 1 - |\rho_{ij}|$, $\rho_{ij}$ being the \emph{Spearman Rank} correlation between the time series of variables $i$ and $j$. 
 Following the previous setting, we perform $1000$ runs over the test set, where in each run, we first randomly pick $c$ clusters and then choose $\mathcal{|S|}$ points from their union. Note that this  method of selection of  $\mathcal{S}$ is the converse of correlated failures. We chose this method since its more direct and is a reasonable proxy for what we want to study.  
We observe in Table \ref{table:conditional_S_delta},  $\Delta_{Ensemble}$ values are higher for lower cluster count (converse for MRR)  as the retrieved neighbors are close w.r.t to the $\mathcal{S}$ variables, which   belong to few clusters. This retrieval generates a higher margin of error in the remaining $\mathcal{N} - \mathcal{S}$ variables and serves as a comparatively worse and biased proxy. In summary, we can conclude that if more uncorrelated variables fail, the error in prediction increases. 

\begin{table}[h]
\small

\caption{Variation of $\Delta_{Ensemble}$ under Correlated Failures setting (on SOLAR). 
Last row shows the MRR values w.r.t optimal nearest neighbor (similar to section \ref{sec:empirical_evaluation}) when $\mathcal{S}$ is sampled from union of given number of clusters.} 
\label{table:conditional_S_delta}

\centering
\setlength\tabcolsep{3.5pt}
\begin{tabular}{|l|l|c|c|c|c|}

\hline
\multirow{2}{0cm}{Forecast  Model} & Error  &  \multicolumn{4}{c|}{\bf \emph{\# Clusters }} \\
\cline{3-6}  

 & & 1 & 4 & 7 & 11 \\
\hline 

\hline
\multirow{2}{*}{ASTGCN} & \emph{MAE}  & 9.42 \% & 5.30 \% & 3.23 \% & 4.32 \% \\
 & \emph{RMSE} & 7.82 \% & 5.45 \% & 3.21 \% & 4.21 \% \\
\hline

\hline
\multirow{2}{*}{MSTGCN} &  \emph{MAE}  & 13.55 \% & 9.96 \% & 8.81 \% & 9.50 \% \\
 & \emph{RMSE} & 13.70 \% & 8.46 \% & 7.02 \% & 7.65 \% \\
\hline

\hline
 & \emph{MRR} & 0.5005 & 0.5706 & 0.5886 & 0.6276  \\
\hline

\end{tabular}

\end{table}


We also analyse the results for the variation of the error metrics over the \emph{Partial} baseline variant, ie variables in $\mathcal{S}$, in table \ref{table:conditional_S}. Contrary to the trend in table \ref{table:conditional_S_delta}, we instead observe increasing MAE and RMSE values since the selection of $\mathcal{S}$ from union of lesser number of clusters imply that the spatial module can model the associations comparatively better.

\begin{table}[h]
\small

\caption{Results for \emph{Partial} baseline variant under Correlated Failures on SOLAR.}
\label{table:conditional_S}

\centering
\setlength\tabcolsep{3.5pt}
\begin{tabular}{|l|l|cccc|}

\hline
\multirow{2}{0cm}{Forecast  Model} & Error &  \multicolumn{4}{c|}{\bf \emph{\# Clusters }} \\
\cline{3-6}  

 & & 1 & 4 & 7 & 11 \\
\hline 

\hline
\multirow{2}{*}{ASTGCN} & \emph{MAE}  & 5.91(2.23) & 5.94(1.66) & 6.04(1.51) & 6.14(1.37) \\
 & \emph{RMSE} & 8.07(3.32) & 8.52(2.76) & 8.75(2.64) & 8.97(2.51) \\
\hline

\hline
\multirow{2}{*}{MSTGCN} & \emph{MAE}  & 4.27(1.59) & 4.39(0.94) & 4.53(0.76) & 4.74(0.73) \\
 & \emph{RMSE} & 6.14(2.35) & 6.24(1.54) & 6.67(1.33) & 7.02(1.44) \\
\hline

\end{tabular}

\end{table}

\section{Conclusion}

In this work, we introduce a new inference task called Variable Subset Forecast in the MTSF setting. We motivate this setting with practical use cases and show that challenging part of the problem is the significantly small sized randomly selected subset at test time.
We propose a simple yet effective and easy to implement solution followed by thorough ablation studies as well as quantitative and qualitative analysis including a comparison against data imputation methods for completeness.




\bibliographystyle{ACM-Reference-Format}
\bibliography{citations}


\newpage 

\appendix

\section{Additional   Experiments }

\subsection{Variation of gap between Oracle and Partial across horizon length}
\label{subsec:gap_subset_oracle_across_horizons}
\begin{figure}[h]
	\begin{subfigure}[b]{0.25\textwidth}
		\centering
		\includegraphics[width=\linewidth]{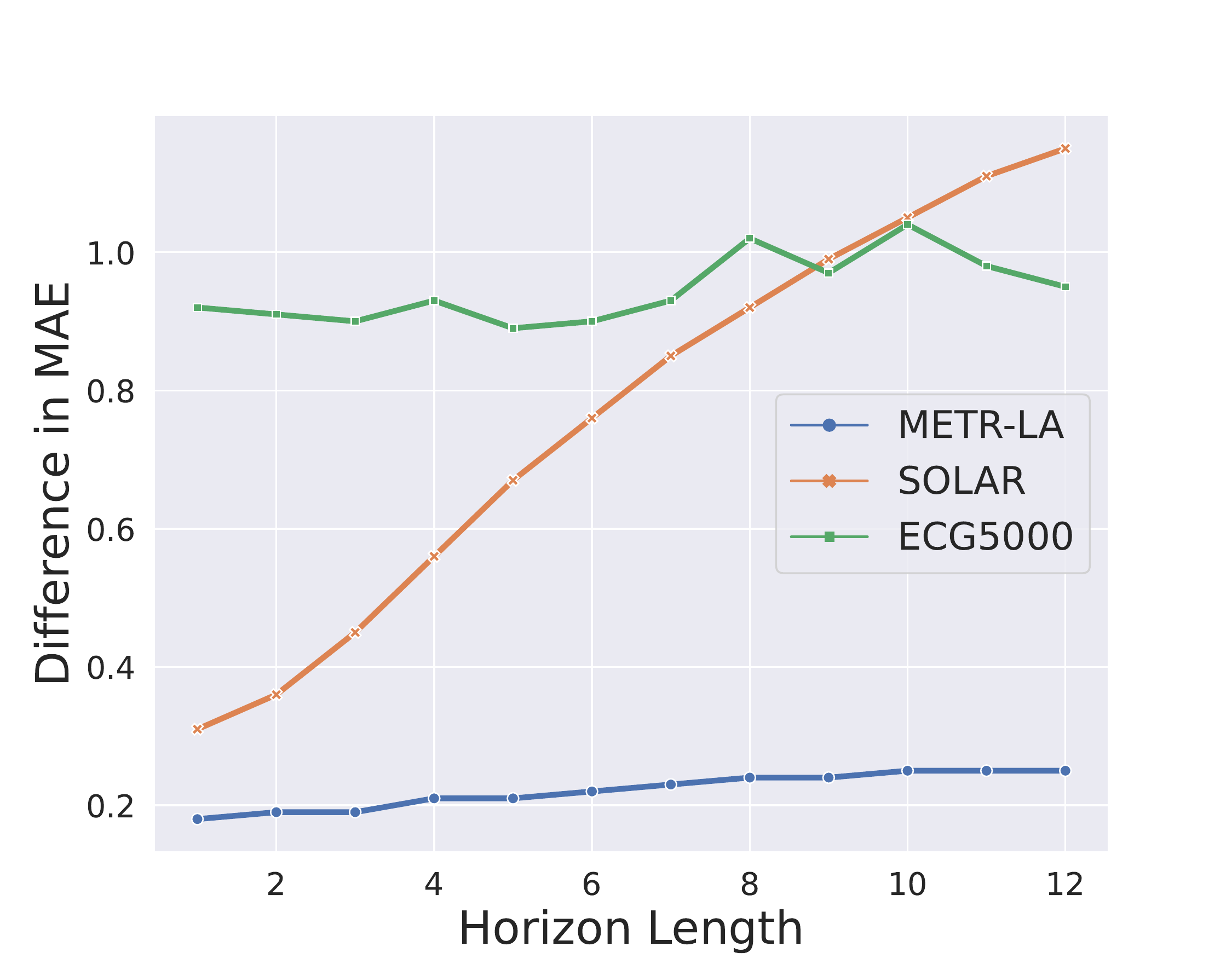}
		\caption{MAE Difference}
		\label{subfig:gap_against_horizon_length_mae}
	\end{subfigure}%
	\begin{subfigure}[b]{0.25\textwidth}
		\centering
		\includegraphics[width=\linewidth]{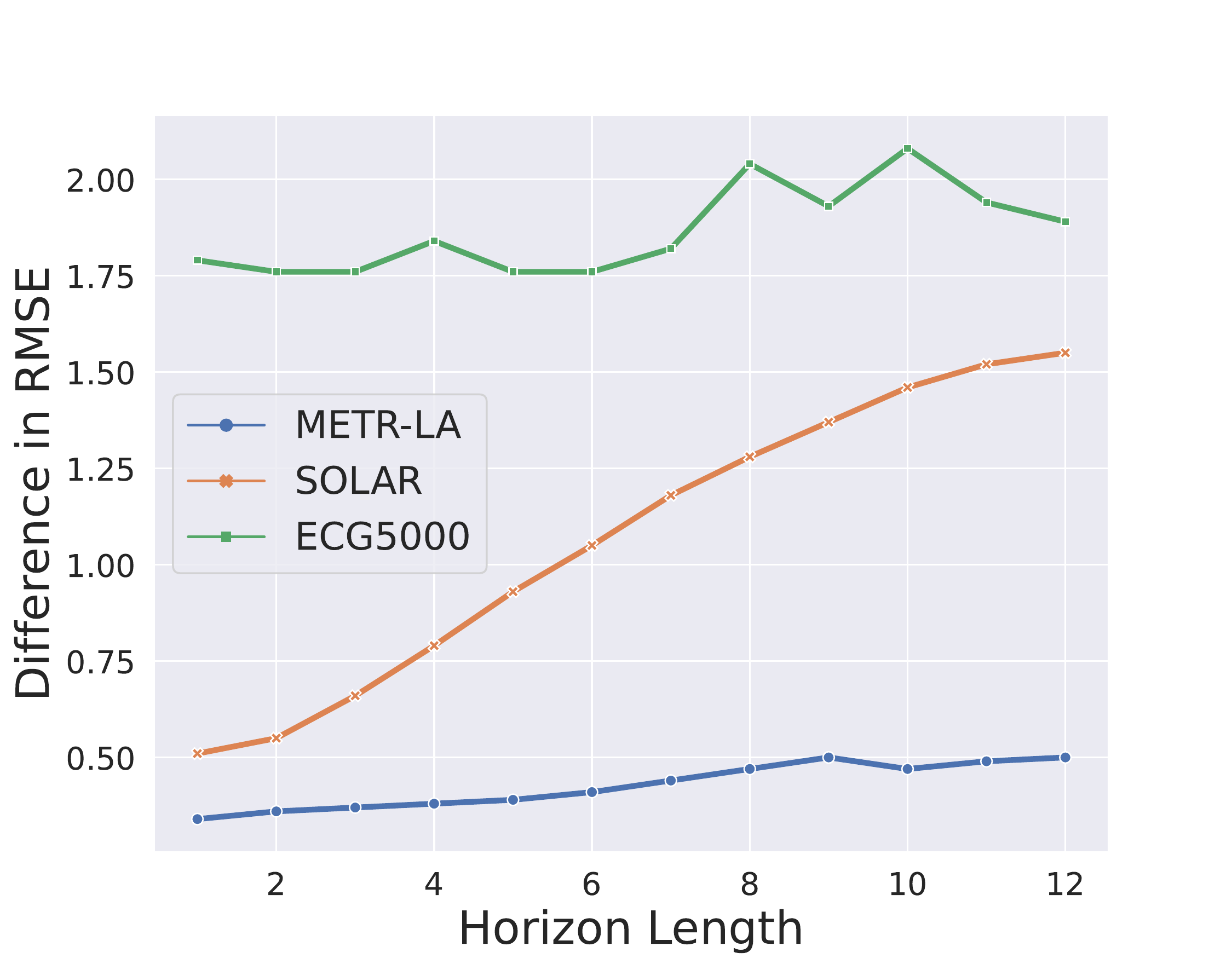}
		\caption{RMSE Difference}
		\label{subfig:gap_against_horizon_length_rmse}
	\end{subfigure}%
	
	\caption{Variation of gap (difference) between Oracle and Partial across horizon length for the error metrics.}	
	\label{fig:gap_against_horizon_length}
\end{figure}

In this section, we study the variation in the \textit{difference} between the error metrics for \emph{Partial} and \emph{Oracle} settings across the horizon length. Figure \ref{fig:gap_against_horizon_length} shows the experiments performed over MSTGCN model. For \emph{METR-LA} and \emph{SOLAR} datasets, this variation is monotonically increasing, but an interesting upward going zig-zag trend surfaces for ECG5000 dataset. We hypothesize that this trend is related to the nature of the temporal patterns. 

From the previous works \cite{wu2020connecting, NEURIPS2020_cdf6581c}, we note that the  errors (MAE, RMSE) increase monotonically (w.r.t ground truth) with horizon length under the standard full variable setting, as it becomes increasingly difficult for the model to accurately forecast values far in future. Following similar reasoning, the \emph{Partial} setting adds another layer of difficulty for the forecast model, thus causing similar deviations from the \emph{Oracle} setting. This also motivates the use of the factor $q$ in the denominator of equation \ref{eq:ensemble_weights_fdw}, as we are computing the forecast differences from a given neighbor $X^{NN_i}$ against the corresponding instance with borrowed data $X^{new_i}$ for all horizons. The discrepancy in the forecasts amplifies over longer horizons.

\subsection{Variable Importance vs $\Delta_{partial}$}
\label{subsec:pagerank_exps} 
 
\begin{figure}[t]
	\begin{subfigure}[b]{0.25\textwidth}
		\centering
		\includegraphics[width=\linewidth]{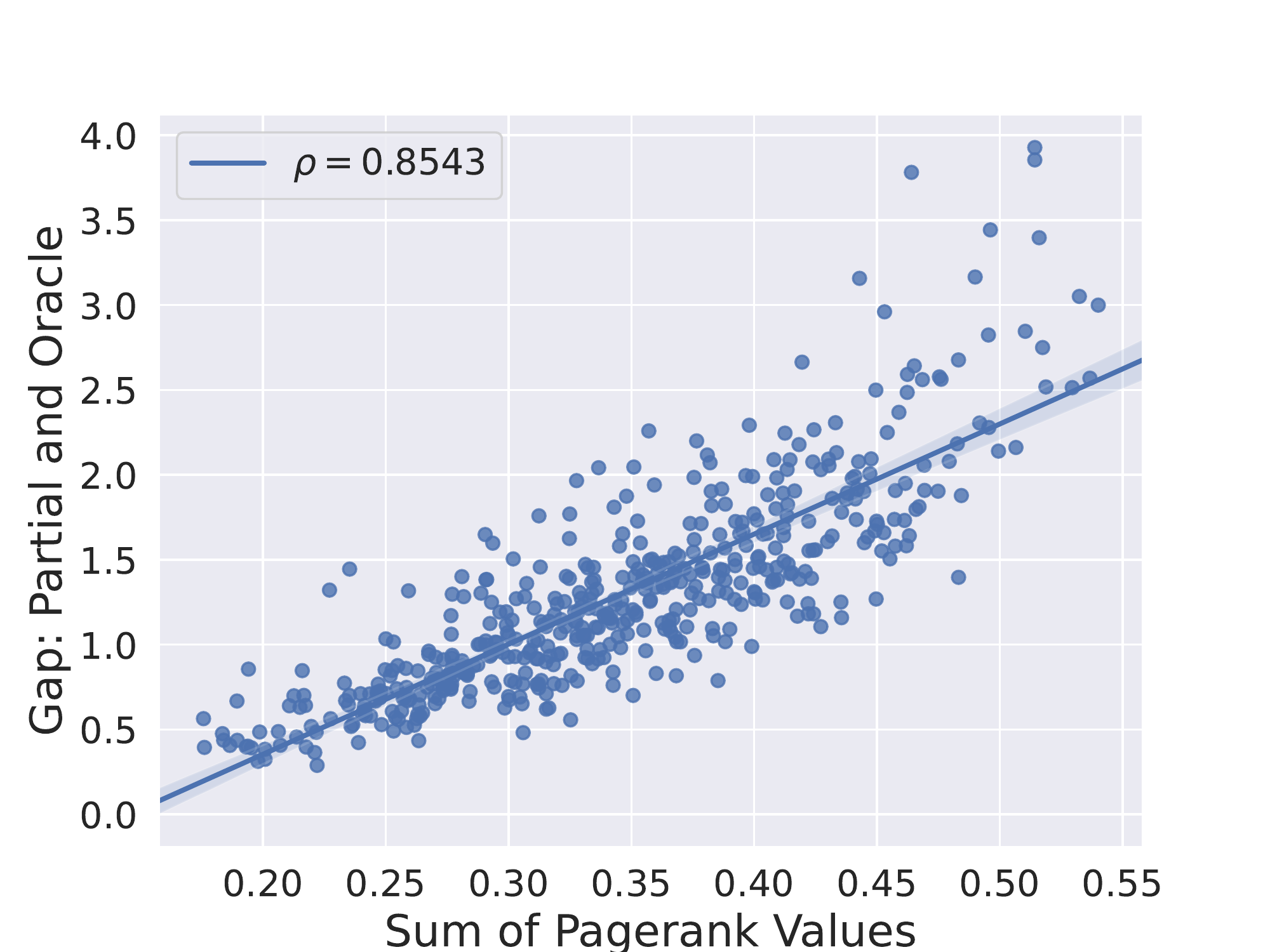}
		\caption{MTGNN on TRAFFIC}
		\label{subfig:pagerank_corrs_mtgnn}
	\end{subfigure}%
	\begin{subfigure}[b]{0.25\textwidth}
		\centering
		\includegraphics[width=\linewidth]{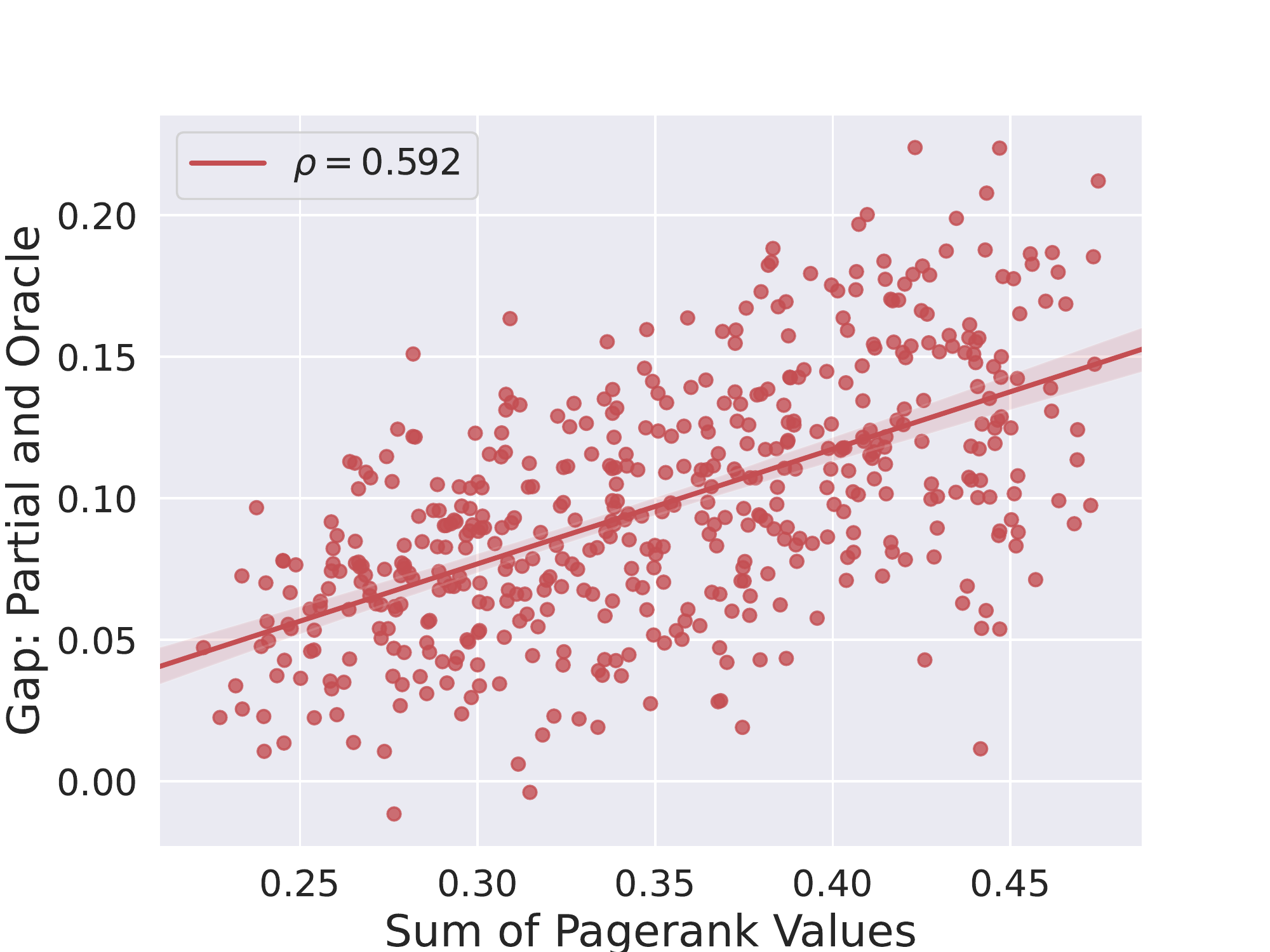}
		\caption{MSTGCN on METR-LA}
		\label{subfig:pagerank_corrs_mstgcn}
	\end{subfigure}%
	
	\caption{Correlation of Variable Importance (measured with \emph{sum of Pagerank} values) to Gap between Partial and Oracle (section \ref{subsec:pagerank_exps}).}	
	\label{fig:pagerank_corrs}
\end{figure}

It is intuitive that for a given model, dataset pair, all variables are not equally important for minimizing the overall MAE wrt ground truth. In this section, we study the effect of variable importance on performance drop of the forecast model. 
We use  \textit{pagerank} to compute variable importance and hence we study only graph based forecast models in this section.  
For a given random selection of $\mathcal{S}$, we compute the sum of page rank values of variables in $\mathcal{N-S}$ and observe the gap between MAE of partial and oracle in $\mathcal{S}$.  Figure~\ref{fig:pagerank_corrs} shows relationship between these two variables where each point represents one random sampling  of $\mathcal{S}$.
We see that there is a very strong correlation ($0.85$ and $0.59$ for the two dataset-model pairs considered) between the two variables. This indicates that removal of important variables (or nodes in graph) has a significant effect on information propagation for the spatial module due to which the forecasts are not well informed and therefore comparatively worse.

\subsection{Hyperparameters}
\label{subsec:hyperparams}
We provide the details of the important hyperparameters used in the experiments in this section. To train the forecast models, we use the optimal values of learning rate, optimizer, batch size, dropout, number of layers etc from the corresponding github implementations. In cases where we do not find the appropriate values, we follow \cite{NEURIPS2020_cdf6581c} and set: learning rate to $1e^{-3}$, dropout to $0.3$, optimizer to \emph{Adam}, batch size to $64$, epochs to $100$, weight decay to $1e^{-4}$, gradient clipping to $5$. We also found \emph{curriculum learning} proposed in \cite{wu2020connecting} to work well while training the models in \emph{Multi-Step} manner \cite{wu2020connecting}. All the models are trained on normalized input, $X^{input} = (X^{input} - \mu)/\sigma$, where $\mu$ and $\sigma$ are scalar \emph{mean} and \emph{standard deviation} of the time series in the training data respectively.
The input and output time series sequence length is fixed to $12$, in order to ensure consistency. Furthermore, as ASTGCN and MSTGCN require a pre-defined graph structure, we use the adjacency matrix learnt by MTGNN model 
as the input graph on some datasets where default adjacency is not available, for both the models. It is evident from the results in Table \ref{table:oracle_vs_subset_setting}, that both ASTGCN and MSTGCN perform reasonably well.\\
The hyperparameters used in our proposed method (section \ref{sec:our_method}) are searched from the sets as follows: exponent $b$ from \{$0.33$, $0.5$, $1$, $2$\}, temperature $\tau$ from \{$0.01$, $0.1$, $1$, $10$\}, number of neighbors from \{$1$, $5$, $10$, $20$, $50$\}. Ablation studies have been done in the main paper. The default values are: $0.5$ for exponent $b$, $0.1$ for temperature $\tau$ and $5$ retrieved neighbors. \\

\begin{table}[h]
\scriptsize

\caption{Dataset Statistics.} 
\label{table:stats_datasets}

\centering
\begin{tabular}{|l|c|c|c|c|c|}

\hline
\cline{1-6}  
Dataset &  \# Timesteps & \# Variables & Sample Rate & Input Length & Output Length \\
\hline
\emph{METR-LA} & 34,272 & 207 & 5 minutes & 12 & 12  \\
\emph{TRAFFIC} & 17,544 & 862 & 1 hour & 12 & 12  \\
\emph{SOLAR}   & 52,560 & 137 & 10 minutes & 12 & 12  \\
\emph{ECG5000} & 5,000  & 140 &     -      & 12  & 12 \\
\hline


\end{tabular}

\end{table}

\section{Scalable k-NN Heuristics for VSF}
Fast Nearest Neighbor Search that scales to billions of datapoints is a topic of active research\cite{johnson2019billion}. For the techniques proposed in this paper, a variation of k-NN retrieval is needed where the similarity needs to be computed on a subset of dimensions.
An optimal solution to this problem is an open research problem by itself and is hence out of scope for this paper. In this section, we present a practical implementation that scales and performs  well.

\subsection{Pre-Processing}
Let $P$ be the forecast time steps that we assume is known during the database construction time and $|\mathcal{T}|$ be the total number of training records (or instances), where each record is a tensor of shape $P \times |\mathcal{N}|$. For ease of notation, we consider $D=1$, however the methodolgy follows for any $D>1$ as well. 
We construct a new table, henceforth called the {\em Query Table} in which each row contains an instance of $\mathcal{T}$, flattened into a vector of shape $P * |\mathcal{N}|$. 
Therefore, the {\em Query Table} will have $|\mathcal{T}|$ rows and $P * |\mathcal{N}|$ columns. Since all the columns are continuous real valued numbers, we   create indexes that are efficient for range queries.


\subsection{Methodology}
Our task remains the same as defined in section \ref{subsec:borrowing_data}: to retrieve $m$ neighbors for the test instance $X^{test}$. We do this by performing  query on the Query Table that has  $F = P * |\mathcal{S}| $ sub-queries. Each sub-query is a range query on a {\em Query Table} column.   For each sub-query $X^{test}_{[p,s]}$ of the tensor $X^{test}$, we find the instance $X^{prime}$ in the {\em Query Table} satisfying the following:
\begin{equation}
    \label{eq:db_search}
    |X^{test}_{[p,s]} - X^{prime}_{[p*|\mathcal{N}| + \sigma(s)]}| \leq b_{\sigma(s)}
\end{equation}
where $\sigma(s)$ is the projection of $s$ to its actual variable in $\mathcal{S}$, $b_{\sigma(s)}$ is a hyperparameter for the corresponding variable. $X^{prime}$ is used as a vector in above equation (note the shape of $X^{prime}$ is $P * |\mathcal{N}|$) with the corresponding element vector indexed by $[p*|\mathcal{N}| + \sigma(s)]$. The set of indices in the table for all such $X^{prime}$ is denoted by $\mathcal{I}({p,s})$. 

Lastly, we obtain the set of desired neighbors by performing intersection of the sets for all elements of $X^{test}$, given by $\mathcal{I} = \bigcap\limits_{p=1}^{P}\bigcap\limits_{s=1}^{|\mathcal{S}|}\mathcal{I}({p,s})$. 

This operation is repeated until we obtain the desired number of $m$ neighbors, thus incurring $\geq 1$ rounds with expansion of $\mathcal{I}$ in each round. In each subsequent round, we multiply some of the $b_{\sigma(s)}$ (out of the total $|\mathcal{S}|$) for which the \emph{L.H.S} of equation \ref{eq:db_search} is tighest w.r.t to $b_{\sigma(s)}$ by a constant factor $u > 1$, which implies loosening the constraints to obtain a larger set.
Since, $|\mathcal{I}| \geq m$, we also perform a linear search over the retrieved neighbors, indexed via $\mathcal{I}$, and retrieve the top-$m$.

\textbf{Selection of Primary Values of $b_{\sigma(s)}$:} We treat $b_{\sigma(s)}$ as hyperparameter for this experiment. Since searching for the value of $b_{\sigma(s)}$ over the real line is difficult, we utilize the training and validation set as follows: (i) for each instance in the validation set, obtain its $\hat{k}^{th}$ neighbor ($\hat{k}$ is a hyperparameter) by using all $\mathcal{N}$ variables, (ii) compute the average distance, w.r.t to the variable dimension, over the $\hat{k}^{th}$ neighbors for all the val set instances. This provides a reasonable proxy to issue a query and perform the search in equation \ref{eq:db_search} for the corresponding $b_{\sigma(s)}$.

\subsection{Experimental Results}

\begin{figure}[h]
    \centering
	\includegraphics[width=\linewidth]{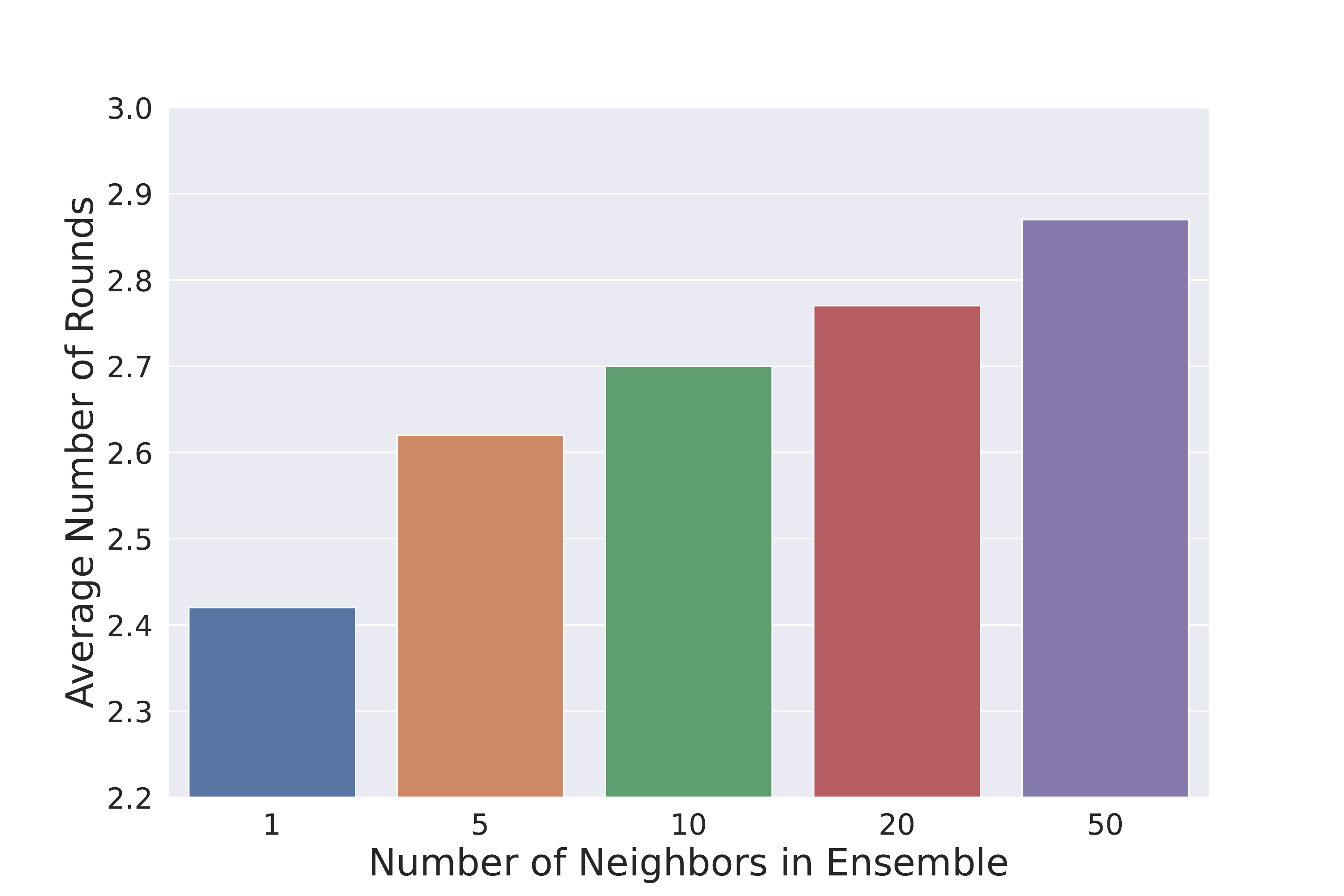}
	
	\caption{Variation of the average number of rounds taken against the desired number of neighbors to fetch. Experimented for MTGNN on ECG5000.}	
	\label{fig:scalability_figs}
\end{figure}

\textbf{Query-Time:}
Figure \ref{fig:scalability_figs} shows the number of rounds required to fetch desired number of neighbors, averaged over $100$ selections of $\mathcal{S}$. As we can see for most of the neighborhood settings experimented in section \ref{subsec:vary_num_neighbors}, we are able retrieve neighbors in at most 3 rounds, with a diminishing increase. The total time incurred is the sum of database search and a linear search over the training instances corresponding to indices in $\mathcal{I}$ (which is a small set), both of which are very fast operations. \\ 

\textbf{Performance Comparison to Direct Neighbor Retrieval (Section \ref{sec:empirical_evaluation}):} We provide a comparison for the values of $\Delta_{Ensemble}$ in the two settings where we retrieve the desired $m$ neighbors via: (i) \emph{Direct} Retrieval (from table \ref{table:ewp_main}) to the neighbors obtained using the approach in this section referred as \emph{Scalable} Retrieval. From the results provided in table \ref{table:scalability_ewp_delta}, its evident that the proposed \emph{scalable search} approach retrieves reasonably good neighbors and can recover performance very close to the \emph{Direct} retrieval. We explicitly call out that the hyperparameters: $b_{\sigma(s)}$, size of $\mathcal{I}$ and the constant multiplication factor $u$ , directly governs the results and thus require careful tuning to provide the desired level of runtime-performance tradeoff.

\begin{table}[h]
\small

\caption{Comparison of $\Delta_{Ensemble}$ for neighbor retrieval using: (i) \emph{Direct} (from table \ref{table:ewp_main}) vs (ii) \emph{Scalable} (proposed in this section) for MTGNN model} 
\label{table:scalability_ewp_delta}

\centering
\setlength\tabcolsep{3.5pt}
\begin{tabular}{|l|rr|rr|}

\hline
   &  \multicolumn{2}{c|}{\bf \emph{SOLAR}}  &  \multicolumn{2}{c|}{\bf \emph{ECG5000}}  \\
\cline{1-5}  
Retrieval &  MAE  & RMSE & MAE & RMSE \\ 
\hline

\hline
\emph{$\Delta_{Ensemble}$ \emph{Direct}} & 30.05 \% & 25.45 \% & 2.14 \% & 3.11 \% \\
\emph{$\Delta_{Ensemble}$ \emph{Scalable}} & 31.28 \% & 26.83 \% & 2.51 \% & 3.29 \% \\
\hline 

\end{tabular}

\end{table}

\end{document}